\documentclass{article}

\PassOptionsToPackage{numbers, compress}{natbib}
\usepackage[dvipsnames]{xcolor}

\definecolor{myblue}{RGB}{31, 119, 180}
\definecolor{myorange}{RGB}{255, 127, 14}
\definecolor{mygreen}{RGB}{44, 160, 44}
\definecolor{myred}{RGB}{214, 39, 40}

\usepackage{iclr2024_conference,times}

\usepackage{thmtools} 
\usepackage{thm-restate}
\usepackage{lipsum}
\usepackage{wrapfig}

\usepackage[utf8]{inputenc} %
\usepackage[T1]{fontenc}    %
\usepackage[allcolors=NavyBlue,colorlinks=true,backref=page]{hyperref}       %
\usepackage{url}            %
\usepackage{tabularx}
\usepackage{booktabs}       %
\usepackage{amsfonts}       %
\usepackage{nicefrac}       %
\usepackage{microtype}      %
\usepackage{xspace}

\usepackage[labelformat=simple,skip=3pt]{subcaption}
\usepackage{tabularray}

\usepackage{wrapfig}

\usepackage[noend]{algpseudocode}

\algrenewcommand\algorithmicrequire{\textbf{Input:}}
\algrenewcommand\algorithmicensure{\textbf{Output:}}

\usepackage{algorithm}

\usepackage{cite}
\usepackage{comment}
\usepackage{prettyref}
\usepackage{amsfonts}
\usepackage{makecell}
\usepackage{adjustbox}
\usepackage{multicol}
\usepackage{multirow}

\usepackage{amsmath}
\usepackage{amssymb}
\usepackage{mathtools}
\usepackage{amsthm}

\usepackage[textsize=scriptsize]{todonotes}
\usepackage{xspace}

\usepackage{titletoc}
\usepackage[toc,page,header]{appendix}
\usepackage{minitoc}

\usepackage{enumitem}
\usepackage{stmaryrd} %

\usepackage{subfiles}

\newcommand\loss{\ell}

\newcommand\Policies{\ensuremath{\Pi}}

\newcommand\ExtendOracleSet{\ensuremath{\Pi^{\mathcal{E}}}}
\newcommand\OracleSet{\ensuremath{\Policies^{\textrm{o}}}}
\newcommand\Policyclass{\ensuremath{\Pi^{\mathcal{L}}}}

\newcommand\policy{\ensuremath{\pi}}

\newcommand\policyN{{\color{blue}\policy_n}}

\newcommand\state{s}
\newcommand\stateDist{d}

\newcommand\transDynamics{\mathcal{P}}
\newcommand\reward{r}
\newcommand\RFunc{\ensuremath{r}}

\newcommand\AFunc{\mathbi{A}} 
\newcommand\Adv{\mathbi{A}}

\newcommand\ENum{\ensuremath{N}}
\newcommand\ONum{\ensuremath{K}}
\newcommand\MDP{\ensuremath{\mathcal{M}}}

\def\mathbi#1{\textbf{\em #1}}

\newcommand\stateSpace{\ensuremath{\mathcal{S}}}
\newcommand\action{\ensuremath{a}}
\newcommand\actionSpace{\ensuremath{\mathcal{A}}}
\newcommand\trajectory{\ensuremath{\mathcal{\tau}}}

\newcommand\horizon{\ensuremath{H}}

\newcommand\VFunc{\ensuremath{\ensuremath{V}}}
\newcommand\QFunc{\ensuremath{\ensuremath{Q}}}
\newcommand\Func{\ensuremath{\ensuremath{f}}}
\newcommand\gradient{\ensuremath{\ensuremath{g}}}

\newcommand\VFuncLearnerN{{\color{blue}\VFunc^{\policy_n}}}

\newcommand\plus{\texttt{+}}

\newcommand\maxplus{\ensuremath{\ensuremath{\textrm{max}^\texttt{+}}}\xspace}
\newcommand\Maxplus{\ensuremath{\ensuremath{\textrm{Max}^\texttt{+}}}\xspace}

\newcommand\maxplusflw{\ensuremath{\ensuremath{\textrm{max}^\texttt{+}}{\textrm{-following}}}\xspace}
\newcommand\Maxplusflw{\ensuremath{\ensuremath{\textrm{Max}^\texttt{+}}{\textrm{-Following}}}\xspace}

\newcommand\maxplusagg{\ensuremath{\ensuremath{\textrm{max}^\texttt{+}}{\textrm{-aggregation}}}\xspace}

\newcommand\polmaxplusflw{\ensuremath{\policy^{\circ}}\xspace}
\newcommand\polmaxplusagg{\ensuremath{\policy^{\circledcirc}}\xspace}

\newcommand\polmaxplusaggAtM{\ensuremath{\policy^{\circledcirc}_m}\xspace}

\newcommand\fplus{\ensuremath{\Func^{\plus}}}
\newcommand\aplus{\ensuremath{\AFunc^{\plus}}}

\newcommand\fplusAtM{\fplus_m}
\newcommand\aplusAtM{\aplus_m}

\newcommand\estfplus{\ensuremath{\hat\Func^{\plus}}}
\newcommand\estaplus{\ensuremath{\hat\AFunc^{\plus}}}

\newcommand\aplusGAE{{\AFunc}^{\textrm{GAE}\texttt{+}}}
\newcommand\estaplusGAE{\hat{\AFunc}^{\textrm{GAE}\texttt{+}}}
\newcommand\estaplusGAEt{\hat{\AFunc}^{\textrm{GAE}\texttt{+}}_t}

\newcommand{\expct}[1]{\mathbb{E}\left[#1\right]}
\newcommand{\expctover}[2]{\mathbb{E}_{#1}\!\left[#2\right]}
\newcommand{\expctoveronly}[1]{\mathbb{E}_{#1}}

\def \argmax {\mathop{\rm arg\,max}}

\newcommand{\algname}{\textsc{RPI}\xspace}

\newif\iffinal

\finaltrue

\iffinal
    \newcommand{\fix}[1]{#1}
    \newcommand{\YC}[1]{}
    \newcommand{\YCinline}[1]{}
    \newcommand{\XL}[1]{}
    \newcommand{\XLinline}[1]{}
    \newcommand{\TY}[1]{}
    \newcommand{\TYinline}[1]{}
    \newcommand{\note}[1]{}
    \newcommand{\pref}[1]{}
    \newcommand{\sout}[1]{}
    \newcommand{\reminder}[1]{#1}
\else
    \newcommand{\fix}[1]{{\color{red} #1}}
    \newcommand{\reminder}[1]{{\color{brown} #1}}
    \newcommand{\YC}[1]{\todo[fancyline,color=NavyBlue!40]{YC: #1}\xspace}
    \newcommand{\YCinline}[1]{\textcolor{NavyBlue}{[YC: #1]}}
    \newcommand{\XL}[1]{\todo[fancyline,color=Maroon!40]{XL: #1}\xspace}
    \newcommand{\XLinline}[1]{\textcolor{Maroon}{[XL: #1]}}
    \newcommand{\TY}[1]{\todo[fancyline,color=pink!40]{TY: #1}\xspace}
    \newcommand{\TYinline}[1]{\textcolor{Maroon}{[TY: #1]}\xspace}

    \newcommand{\note}[1]{{\color{purple}[XL: #1]}}
    
    \newcommand{\pref}[1]{{\color{blue}(\ref{#1})}}
    \usepackage[normalem]{ulem}
\fi

\newcommand{\defref}[1]{Definition~\ref{#1}}
\newcommand{\tabref}[1]{Table~\ref{#1}}
\newcommand{\figref}[1]{Fig.~\ref{#1}}
\newcommand{\eqnref}[1]{\text{Eqn.}~\ref{#1}}

\newcommand{\appref}[1]{Appendix~\ref{#1}}

\newcommand{\corref}[1]{Corollary~\ref{#1}}
\newcommand{\propref}[1]{Proposition~\ref{#1}}
\newcommand{\lemref}[1]{Lemma~\ref{#1}}

\newcommand{\algoref}[1]{Algorithm~\ref{#1}}
\newcommand{\remarkref}[1]{Remark~\ref{#1}}

\newcommand{\paren} [1] {\ensuremath{ \left( {#1} \right) }}

\newcommand{\bracket}[1]{\left[#1\right]}

\newcommand{\curlybracket}[1]{\ensuremath{\left\{#1\right\}}}

\theoremstyle{plain}
\newtheorem{theorem}{Theorem}[section]
\newtheorem{proposition}[theorem]{Proposition}
\newtheorem{lemma}[theorem]{Lemma}
\newtheorem{corollary}[theorem]{Corollary}
\theoremstyle{definition}
\newtheorem{definition}[theorem]{Definition}

\newtheorem{remark}[theorem]{Remark}

\newcommand\RewardFunc{\ensuremath{r}}
\newcommand\var{\ensuremath{{v}}}
\newcommand\threshold{\ensuremath{{\Gamma_{\state}}}}

\title{Blending Imitation and Reinforcement Learning for Robust Policy Improvement}

\author{Xuefeng Liu\textsuperscript{1}\thanks{Correspondence to: Xuefeng Liu <\href{mailto:xuefeng@uchicago.edu}{xuefeng@uchicago.edu}>.} ,~\textbf{Takuma Yoneda\textsuperscript{2}},~\textbf{Rick L. Stevens\textsuperscript{1}},~\textbf{Matthew R.\ Walter\textsuperscript{2}},~\textbf{Yuxin Chen\textsuperscript{1}} \\
\textsuperscript{1}Department of Computer Science, University of Chicago\\
\textsuperscript{2}Toyota Technological Institute at Chicago 
}

\begin{document}

\iclrfinalcopy
\maketitle
\doparttoc %
\faketableofcontents %

\begin{abstract}
While reinforcement learning~(RL) has shown promising performance, its sample complexity continues to be a substantial hurdle, restricting its broader application across a variety of domains. 
Imitation learning~(IL) utilizes oracles to improve sample efficiency, yet it is often constrained by the quality of the oracles deployed. %
To address the demand for robust policy improvement in real-world scenarios, we introduce a novel algorithm, Robust Policy Improvement (\algname), %
which actively interleaves between IL and RL based on an online estimate of their performance. %
\algname draws on the strengths of IL, using oracle queries to facilitate exploration---an aspect that is notably challenging in sparse-reward RL---particularly during the early stages of learning. As learning unfolds, \algname gradually transitions to 
RL, effectively treating the learned policy as an improved oracle. %
This algorithm is capable of learning from and improving upon a diverse set of %
black-box oracles. %
Integral to \algname are Robust Active Policy Selection~(RAPS) and Robust Policy Gradient~(RPG), both of which reason over whether to perform state-wise imitation from the oracles or learn from its own value function %
when the learner's performance surpasses that of the oracles in a specific state. 
Empirical evaluations and theoretical analysis validate that \algname excels in comparison to existing state-of-the-art methods, showing superior performance across various domains.
Please checkout our website\footnote{https://robust-policy-improvement.github.io/}.

\end{abstract} 

\section{Introduction}

Reinforcement learning (RL) has shown significant advancements, surpassing human capabilities in diverse domains such as Go~\citep{silver2017mastering}, video games~\citep{berner2019dota, mnih2013playing}, and Poker~\citep{zhao2022alphaholdem}. 
Despite such achievements, the application of RL is largely constrained by its substantially high sample complexity, 
particularly in fields like robotics~\citep{singh2022reinforcement} 
and healthcare~\citep{han2023survey}, where the extensive online interaction for trial and error is often impractical. %

Imitation learning (IL)~\citep{osa18} improves sample efficiency by allowing the agent to replace
some or all environment interactions with demonstrations provided by an oracle policy.
The efficacy of IL heavily relies on access to near-optimal oracles for approaches like behavior cloning \citep{pomerleau1988alvinn, zhang18a} or inverse reinforcement learning~\citep{abbeel04, finn16a, ho16, ziebart08}. Interactive IL techniques, such as DAgger~\citep{ross2011reduction} and AggreVate(D)~\citep{ross2014reinforcement, sun2017deeply},
similarly assume that the policy we train (i.e., \emph{learner} policy) can obtain demonstrations from a near-optimal oracle.
When we have access to rewards, the learner has the potential to improve and outperform the oracle.
THOR~\citep{sun2018truncated} exemplifies this capability by utilizing a near-optimal oracle for cost shaping, optimizing the $k$-step advantage relative to the oracle's value function (referred to as ``cost-to-go oracle'').

However, in realistic settings, obtaining optimal or near-optimal oracles is often infeasible. Typically, learners have access to \textit{suboptimal} and \textit{black-box} oracles that may not offer optimal trajectories or quantitative performance measures in varying states, requiring substantial environment interactions to identify state-wise optimality. 
Recent approaches, including LOKI~\citep{cheng2018fast} and TGRL~\citep{shenfeld2023tgrl} aim to tackle this issue by combining IL and RL.
They focus on a single-oracle setting, whereas MAMBA~\citep{cheng2020policy} and MAPS~\citep{liu2023active} learn from multiple oracles.
These approaches demonstrate some success, but often operate under the assumption that at least one oracle provides optimal actions in any given state, which does not always hold in practice. %
In situations where no oracle offers beneficial advice for a specific state, it is more effective to learn based on direct reward feedback. Our work intend to bridge this gap by adaptively blending IL and RL in a unified framework.

\paragraph{Our contributions.}%
In this paper,
we present \maxplus, %
a learning framework devised to enable robust learning in unknown Markov decision processes (MDP) by interleaving RL and IL, leveraging multiple suboptimal, black-box oracles. 
Within this framework, we introduce \textit{Robust Policy Improvement} (\algname), a novel policy gradient algorithm designed to facilitate learning from a set of black-box oracles. 
\algname comprises two innovative components:
\begin{enumerate}
    \item \textit{Robust Active Policy Selection} (RAPS), improving value function estimators of black-box oracles efficiently, and
    \item \textit{Robust Policy Gradient} (RPG), executing policy gradient updates within an actor-critic framework based on a newly devised advantage function.
\end{enumerate}
Our algorithm strikes a balance between learning from these suboptimal oracles %
and self improvement through active exploration in states where the learner has surpassed the oracle's performance.
We provide a theoretical analysis of our proposed method, %
proving that it ensures %
a performance lower bound no worse than that of the competing baseline \citep{cheng2020policy}.
Through extensive empirical evaluations on eight %
different tasks from DeepMind Control Suite \citep{tassa2018deepmind} and Meta-World \citep{yu2020meta}, we empirically demonstrate that \algname outperforms contemporary methods and then ablate its core components. %

\section{Related Work}

\paragraph{Online selection of suboptimal experts.}%
CAMS~\citep{liucontextual,liu2022cost} learns from multiple suboptimal black-box experts to perform model selection based on a given context,
but is only applicable in stateless online learning settings. 
Meanwhile, SAC-X~\citep{riedmiller2018learning} learns the intention policies (oracles), each of which optimizes their own auxiliary reward function, and then reasons over which of these oracles to execute as a form of curriculum learning for the task policy. 
LfGP~\citep{ablett2023learning} combines adversarial IL with SAC-X to improve exploration. 
Defining auxiliary rewards requires the task to be decomposed into smaller subtasks, which may not be trivial. %
Further, they query the intention policies several times within a single episode.
Unlike CAMS and SAC-X, which rely on selecting expert policies to perform sub-tasks, our approach trains an independent learner policy. It acquires expertise from sub-optimal experts using only a single oracle query per episode, thus having the potential to surpass these oracles through global exploration. %

\looseness -1 \paragraph{Policy improvement with multiple experts.}
Recent works attempt to learn from suboptimal black-box oracles while also utilizing rewards observed under the learner's policy. 
\fix{Active offline policy selection (A-OPS) ~\citep{konyushova2021active} utilizes policy similarities to enhance value predictions. 
However, A-OPS lacks a learner policy to acquire expertise from these offline policies. ILEED \citep{ilbelaiev} distinguishes between oracles based on their expertise at each state but is constrained to pure offline IL settings.}
InfoGAIL~\citep{li2017infogail} conditions the learned policy on latent factors that motivate demonstrations of different oracles. OIL~\citep{li2018oil} tries to identify and follow the best oracle in a given situation. 
SFQL~\citep{barreto2017successor} proposes \textit{generalized policy improvement} with successor features. MAMBA~\citep{cheng2020policy} utilizes an advantage function with geometric weighted generalization and achieves a larger policy improvement over SFQL\fix{, while addressing the above two important questions with theoretical support}. %
MAPS~\citep{liu2023active} improves on the sample efficiency and performance of MAMBA by proposing active policy selection and state exploration.
However, \fix{even when the quality of the oracle set} is poor, %
these algorithms will still resort to imitation learning with the inferior oracles. In contrast, our algorithm performs self-improvement, employing imitation learning only on states for which an oracle outperforms the learner.

\begin{table*}[H]
\scalebox{0.4}{
\begin{tabular}{l l l l l l l l l l l l l}
\toprule
\textbf{Algorithm}
& \textbf{\makecell[l]{BEHAVIOR \\ CLONING}}
& \textbf{SMILE(10)}
& \textbf{SEARN}
& \textbf{DAGGER(11)}
& \textbf{AGGREVATE(14)}
& \textbf{GAE(15)}
& \textbf{A-OPS(21)}
& \textbf{LEAQI(20)}
& \textbf{MAMBA(20)} 
& \textbf{ILEED(22)}
& \textbf{CAMS(22)}
& \textbf{\algname}
\\
& \makecell[l]{\citep{pomerleau1988alvinn}}
& \citep{ross2010efficient}
& \citep{daume2009search}
& {\citep{ross2011reduction}}
& {\citep{ross2014reinforcement}} 
& {\citep{schulman2015high}}
& {\citep{konyushova2021active}}
& {\citep{brantley2020active}} 
& {\citep{cheng2020policy}} 
& {\citep{ilbelaiev}}
& {\citep{liu2022cost}}
& {(ours)}
\\
\hline
criterion 
& offline IL 
& offline IL
& offline IL
& online IL
& online IL
& RL
& policy selection
& interactive IL
& online IL/RL
& offline IL
& active model selection
& online IL/RL \\
\hline
information  
& full
& full
& full 
& full
& full
& -
& partial / full
& full
& full
& partial / full
& full
& partial / full\\ 
\hline
stateful
& yes
& yes
& yes
& yes
& yes 
& -
& -
& yes 
& yes
& yes 
& no
& yes \\
\hline
active
& no 
& no 
& no 
& no
& no 
& -
& -
& yes
& no
& no
& yes
& yes \\
\hline
multiple experts  
& no
& no 
& no 
& no 
& no 
& -
& yes
& no 
& yes
& yes
& yes 
& yes \\
\bottomrule
\end{tabular}
}
\caption{Algorithms Characteristics}
\label{table:alg_characteristics}
\end{table*}

\section{Preliminaries}

We consider a finite-horizon Markov decision process (MDP)  $\MDP_0=\langle\stateSpace,\actionSpace,\transDynamics,\RFunc, \horizon\rangle$ with state space $\stateSpace$, action space $\actionSpace$, unknown stochastic transition dynamics $\transDynamics: \stateSpace \times \actionSpace \rightarrow \Delta(\stateSpace)$, unknown reward function $\RFunc: \stateSpace \times \actionSpace \rightarrow [0,1]$, and  {episode} horizon $H$. We 
define total number of \fix{training steps (rounds)}
as $\ENum$ and assume access to a (possibly empty) set of $\ONum$ oracles, defined as $\Policies=\{\policy^k\}_{k=1}^{\ONum}$, where $\policy_k: \stateSpace \rightarrow \Delta(\actionSpace)$.
The \textit{generalized Q-function} with respect to a general function $ {\Func}:\stateSpace\rightarrow \mathbb{R}$ is defined as: 
\begin{equation*}
\QFunc^{\Func}\paren{\state,\action}:=\reward\paren{\state,\action}+\expctover{\state'\sim \transDynamics|\state,\action}{\Func\paren{\state'}}. %
\end{equation*}
When {$\Func(\state)$} is the value function of a particular policy $\pi$, the generalized Q-function can be used to recover the policy's Q-function $Q^\pi(\state,\action)$. 
We denote the \textit{generalized advantage function}
with respect to $\Func$ as 
\begin{align*}
    \AFunc^{\Func}\paren{\state,\action} &= \QFunc^{\Func}\paren{\state,\action}-\Func\paren{\state} =\RFunc\paren{\state,\action}+\expctover{\state'\sim\transDynamics|\state,\action}{\Func\paren{\state'}}-\Func\paren{\state}.    
\end{align*}
Given an initial state distribution $\stateDist_0 \in \Delta(\stateSpace)$, let $\stateDist_t^{\policy}$ denote the distribution over states at time $t$ under policy $\pi$. The state visitation distribution under $\pi$ can be expressed as $\stateDist^{\policy}:= \frac{1}{H}\sum_{t=0}^{H-1} \stateDist_t^{\policy}$. The value function of the policy $\pi$ under $d_0$ is denoted as:
\begin{align*}
    \begin{split}
        \VFunc^{\policy}\paren{\stateDist_0}&=\expctover{\state_0\sim \stateDist_0}{\VFunc^{\policy}\paren{\state}} = \expctover{\state_0 \sim \stateDist_0}{\expctover{\trajectory_0\sim \rho^{\policy}|\state_0}{\sum_{t=0}^{H-1}\reward\paren{\state_t,\action_t}}}
    \end{split}
\end{align*}
where $\rho^\pi(\tau_t \mid s_t)$ is the distribution over trajectories $\tau_t = \{\state_t, \action_t, \ldots, \state_{\horizon-1}, \action_{\horizon-1}\}$
under policy $\pi$. %
The goal is to find a policy $\policy=\argmax_{\pi}J\paren{\pi}$ maximizing the expected return
\begin{equation}\label{eq:expect_value}
    J\paren{\policy} = \expctoveronly{\state\sim\stateDist_0} \bracket{\VFunc^{\policy}\paren{\state}}.
\end{equation}

\section{Policy Improvement with Perfect Knowledge of Oracle Set}

We now present a reinforcement learning framework in the presence of an imitation learning oracle set, \fix{which is inspired from \citet{cheng2020policy,liu2023active}}. In this section, we assume that we have perfect knowledge of the underlying MDP and each oracle's value function. We will relax these assumptions in the next section. 

\paragraph{Max-following.}  Given a collection of $k$ imitation learning \textit{oracles} $\OracleSet=\{\policy^k\}_{k\in\bracket{\ONum}}$, 
the \textit{max-following} policy 
is a greedy policy that selects the oracle with the highest expertise in any given state.
 The {max-following} policy 
is sensitive to the {quality}
of the oracles. Specifically, if all oracles perform worse than the learner policy at a given state, the {max-following policy}
will still naively imitate the best (but poor) oracle. Instead, it would be more prudent to follow the learner's guidance in these cases.
\begin{definition}\label{def:extended_oracle_set}
    {(\textbf{Extended Oracle Set})}. Let $\OracleSet=\{\policy^k\}_{k\in\bracket{\ONum}}$ be the given black-box oracle set, 
    {$\Policyclass=\curlybracket{\policy_n}_{n\in \bracket\ENum}$} be the learner's policy class, where $\policyN$ denotes that the policy has been updated for $n$ rounds.
    We define the \emph{extended oracle set} at the $n$-th round as 
    \begin{equation}
        \ExtendOracleSet=\OracleSet \cup \curlybracket{\policyN} = \curlybracket{\policy^1, \ldots, %
        \policy^\ONum
        ,\policyN}
        .
    \end{equation}
\end{definition}
\begin{remark}
The learner policy in the extended oracle set is updated at each round.
\end{remark}

\subsection{\Maxplus Aggregation}
Based on the extended oracle set, we first introduce the advantage function $\aplus$ and the baseline value function $\fplus$ as follows:
\begin{definition}{(\textbf{$\aplus$ Advantage Function})}.
    Given $k$ oracles $\pi^1, \ldots, \pi^k$ and the learner policy $\policyN$, we define $\aplus$ advantage function as :
    \begin{equation}\label{eq:A_plus_advantage_function}
        \aplus\paren{\state,\action} :=  %
        \reward\paren{\state,\action} + \expctover{\state' \sim \transDynamics \vert \policy, \state}{\fplus\paren{\state'}} - \fplus\paren{\state},
    \end{equation}
{where $\fplus\paren{\state}$ is the baseline value function, 
defined as:}
    \begin{equation}\label{eq:f_plus}
    \fplus\paren{\state} = \max_{k\in{\bracket{\lvert\ExtendOracleSet\rvert}}} \VFunc^k\paren{\state}, \textrm{ where } \bracket{\VFunc^k}_{k\in\bracket{\lvert \ExtendOracleSet \rvert}} := \bracket{\VFunc^{\policy^1}, \ldots, \VFunc^{\policy^{\ONum}},\VFuncLearnerN}.
    \end{equation}
\end{definition}
$\fplus\paren{\state}$ focuses exclusively on optimizing oracle selection for a single state, assuming that the selected policy will be followed for the remainder of the trajectory. To optimize the oracle selection
for every encountered state, we introduce the \maxplusflw policy, which acts as a greedy policy, adhering to the optimal policy within the \emph{extended} oracle set for any given state.
\begin{definition}
{(\textbf{\Maxplusflw Policy})}. \looseness-1 Given extended oracle set $\ExtendOracleSet$, the \emph{\maxplus-following}
    {policy}
    \begin{equation}\label{eq:max_plus_following}
        \polmaxplusflw\paren{\action \mid \state} := \policy^{k^{\star}}\paren{\action \mid \state}, \textrm{ where } 
        k^{\star} := \argmax_{k\in \bracket{\lvert \ExtendOracleSet \rvert}} \, \VFunc^k\paren{\state}, \lvert \ExtendOracleSet\rvert = \ONum + 1,  \VFunc^{\ONum+1} = \VFuncLearnerN.
    \end{equation}
\end{definition}

\begin{proposition}\label{pro:policy_oplus}
Following $\polmaxplusflw$ is as good or better than imitating the single-best policy in $\ExtendOracleSet$.%
\end{proposition}
With a slight abuse of notation, we use $\aplus\paren{\state,\polmaxplusflw}$ to denote the generalized advantage function of the policy $\polmaxplusflw$ at $\state$. As proved in the Appendix, the function $\aplus\paren{\state,\polmaxplusflw}\geq 0$ (\appref{pr:advantage}) and that the value function for the \maxplusflw policy satisfies $\VFunc^{\polmaxplusflw}(\state) \geq \fplus\paren{\state} = \max_{k\in\bracket{|\ExtendOracleSet|}}\VFunc^{k}(\state)$ (\appref{lem:PDL}). %
{This indicates that following $\polmaxplusflw$ is at least as good as or better than imitating a single best policy in $\ExtendOracleSet$.}
Thus, $\polmaxplusflw$ is a {valid} approach
to robust policy learning in the multiple oracle setting.
The \maxplusflw 
policy $\polmaxplusflw$ is better than the max-following policy
when the learner's policy is better than any oracle for a given state.
On the other hand, when the value of a specific oracle $\VFunc^k$
is always better than all other policies for all states, $\polmaxplusflw$ simply reduces to the corresponding oracle $\policy^k$. %
 {This is not ideal because the value $\VFunc^k(s)$ assumes to keep rolling out the same oracle $\policy^k$ from state $s$ until  termination, without making improvement by looking one step ahead and searching for a better action.}
To address this,
we propose the \textit{\maxplus-aggregation} policy as follows.

\begin{definition}(\textbf{\Maxplus-Aggregation Policy}\footnote{\fix{When we exclude the learner's policy from the extended oracle set, this reduces to the $\max$-aggregation policy, which was used in MAMBA~\citep{cheng2020policy}.}}).\label{def:maxplusagg}
    For state $\state$, the \maxplusagg policy $\polmaxplusagg$ performs one-step improvement {and takes the action with largest advantage over $\fplus$},
    \begin{equation}\label{eq:max_plus_aggregation}
        \polmaxplusagg\paren{\action \mid \state}=\delta_{\action=\action^{\star}}, \textrm{where } \action^{\star}= \argmax_{a\in\actionSpace} \aplus\paren{\state, \action} \textrm{and $\delta$ is the Dirac delta distribution.} 
    \end{equation}
\end{definition}
Although the \maxplusflw policy $\polmaxplusflw$ improves upon the {max-following} policy,
it does not perform self-improvement. 
In contrast, the \maxplusagg policy $\polmaxplusagg$ looks one step ahead and {makes the largest one-step advantage improvement} with respect to $\fplus$. {Thus,} in the degenerate case  {where %
\polmaxplusflw is equivalent to the single best policy},
$\polmaxplusagg$ 
outperforms the best single policy in
$\ExtendOracleSet$ for all states. 
Since $\aplus(\state,\polmaxplusagg) \geq \aplus\paren{\state,\polmaxplusflw} \geq 0$ 
for any state $\state$ by \corref{col:performance_improvable} and \propref{pro:policy_oplus}, we conclude that the \maxplusagg policy $\polmaxplusagg$ is a suitable policy benchmark for the robust policy learning setting as well. 
We note that the baseline $\fplus(\state)$ corresponds to the value of choosing the single-best policy in $\ExtendOracleSet$ at state $\state$ %
{and rolling it out throughout the rest of the %
episode.} %
In contrast, $\polmaxplusflw$ and $\polmaxplusagg$ %
 {optimize the oracle selection}
at every remaining step in the trajectory. %
This work is therefore built on $\polmaxplusagg$.

\begin{remark}\textbf{(Empty Oracle Set)} Up to this point, we have primarily assumed a non-empty oracle set {$\OracleSet$} and an extended oracle set of size $\lvert \ExtendOracleSet \rvert \geq 2$. Given an empty oracle set {$\OracleSet$}, $\ExtendOracleSet$ will only contain the learner policy. In this case, $\fplus \equiv \VFunc^{\polmaxplusflw}$ and $\polmaxplusflw$ will not improve,
 while $\polmaxplusagg$ reduces to pure reinforcement learning, performing self-improvement by using the advantage function $\aplus$. 
\end{remark}

\vspace{-5mm}
\section{
Robust 
{Policy Improvement with Black-Box Oracle Set}}

\vspace{-2mm}
Improving a policy from the  {\maxplus~baseline $\fplus$} (Eqn.~\ref{eq:f_plus}) %
is the key to learning robustly via IL and RL. This requires knowledge of the MDP and the oracles' value functions, however,
the oracles are presented to the learner as black-box policies with unknown value functions.

A critical challenge to use $\fplus\paren{\state} = \max_{k\in\bracket{\lvert \ExtendOracleSet \rvert}} \VFunc^{k}\paren{\state}$ as a baseline is that it changes as training goes,  %
\fix{whereas MAMBA assumes a static baseline function.}
In the following analysis we resort to a slightly weaker baseline, $\fplusAtM := \max_{k\in\bracket{\lvert \OracleSet \cup \{\policy_m\}\rvert}} \VFunc^{k}\paren{\state}$, where $m \ll N$ is an intermediate step in the learning process, and $N$ is the total number of rounds. %
Similarly, we define $        \aplusAtM\paren{\state,\action} :=  %
        \reward\paren{\state,\action} + \expctover{\state' \sim \transDynamics \vert \policy, \state}{\fplusAtM\paren{\state'}} - \fplusAtM\paren{\state},
$ as the corresponding advantage function, and $\polmaxplusaggAtM$ as the corresponding \maxplusagg policy by setting $\aplus = \aplusAtM$ in \defref{def:maxplusagg}. 
In the following, we use the baseline value $\fplusAtM$, 
and reformulate the problem in an online learning setting ~\citep{ross2011reduction, ross2014reinforcement, sun2017deeply, cheng2020policy, liu2023active} for black-box oracles. Following MAMBA's analysis, we first assume that the oracle value functions are known but the MDP is unknown, followed by the case that the value functions are unknown.

\paragraph{Unknown MDP with known value functions.} If the MDP is unknown,  %
we can regard $\stateDist^{\policy_n}$ as an adversary in online learning and establish the online loss for  
round
$n$ as
\begin{equation}\label{eq:max_plus_loss}
    \loss_{n}\paren{\policy} := -\horizon
    \mathbb{E}_{\state\sim\stateDist^{\policy_n}} { {{\mathbb{E}}_{\action\sim {\policy|\state}}}}
    \bracket{\aplus\paren{\state, {\action}}}.
\end{equation}
\lemref{lem:PDL} and \propref{pro:policy_oplus} suggest that making $\loss_{n}\paren{\policy}$ small ensures that $\VFunc^{\policy_n}\paren{\stateDist_0}$ {achieves better
performance than}
$\fplusAtM\paren{\stateDist_0}$ for $m < n$. Averaging over $\ENum$ 
rounds
of online learning, we obtain
\begin{equation}\label{eq:v_f}
    \frac{1}{N}\sum_{n\in\bracket{\ENum}}\VFunc^{\policy_n}\paren{\stateDist_0} = \fplusAtM\paren{\stateDist_0} + \Delta_N -\epsilon_N\paren{\Policyclass}- {\textrm{Regret}^{\mathcal{L}}_\ENum},
\end{equation}
where 
$\textrm{Regret}^{\mathcal{L}}_\ENum:= {\frac{1}{\ENum}}{(\sum_{n=1}^{\ENum}\loss_{n}\paren{\policy_n}-\min_{\policy\in \Policyclass}\sum_{n=1}^{\ENum}\loss_n\paren{\policy})}$ depends the learning speed of an online algorithm, $\Delta_{\ENum} := -\frac{1}{N}\sum_{n=1}^{\ENum} \loss_n(\polmaxplusaggAtM)$ is the loss of the baseline \maxplusagg policy \polmaxplusaggAtM, and 
$\epsilon_N(\Policyclass):= \min_{\policy\in\Policyclass}\frac{1}{\ENum}(\sum_{n=1}^N\loss_n(\policy)-\sum_{n=1}^{\ENum}\loss_n(\polmaxplusaggAtM))$ %
expresses the quality of oracle class, {where $\Policyclass$ is specified in \defref{def:extended_oracle_set}}. If $\polmaxplusaggAtM \in \Policyclass$, we have $\epsilon_N(\Policyclass) = 0$. 
Otherwise, %
$\epsilon_{\ENum}(\Policyclass) > 0$. By \propref{pro:policy_oplus}, $\aplus(\state,\polmaxplusaggAtM) \geq 0$ and, in turn, $\Delta_N\geq 0$. 
If $\polmaxplusagg\in \Policyclass$, using a no-regret algorithm to address this online learning problem will produce a policy that achieves performance of at least %
$\expctoveronly{\state\sim \stateDist_0}[\fplusAtM\paren{\state}]+\Delta_{N}+O(1)$ after $N$ iterations.

\begin{figure*}[t]
\vspace{-5mm}
    \begin{subfigure}{1\textwidth}
        \centering
        \includegraphics[%
        width=8cm,  clip={0,0,0,0}]{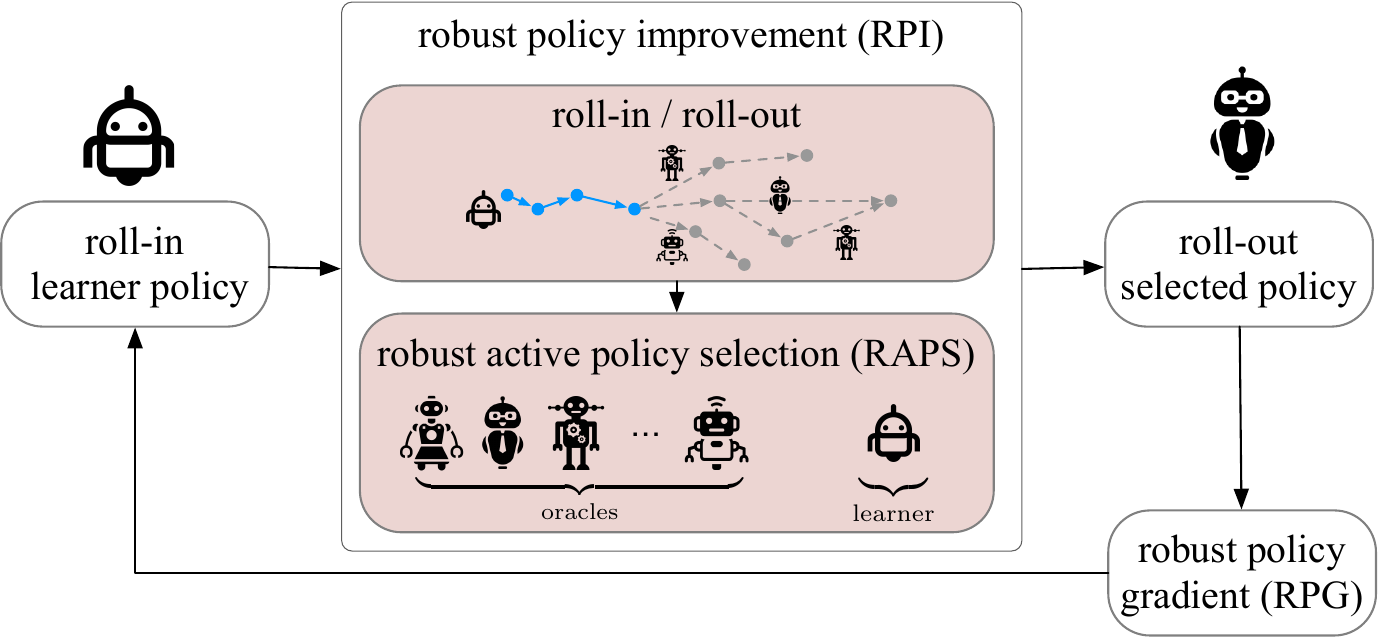}
    \end{subfigure}\hfil
    \caption{ Method Overview.}\label{fig:ucbvsmean}
        \vspace{-5mm}
\end{figure*}

\paragraph{Unknown MDP with unknown value function.} In practice, the value functions of the oracle set %
are unavailable. $\fplus$ and $\aplus$ need to be approximated by $\estfplus$ and $\estaplus$.  
We compute the sample estimate of the gradient as follows:
\begin{equation}\label{eq:max_plus_gradient}
    \nabla \hat{\loss}_n\paren{\policy_n} = - \horizon\expctoveronly{\state\sim\stateDist^{\policy_n}}\expctover{\action \sim \policy_n \vert \state}{\nabla \log{\policy_n\paren{\action \mid \state}}~\estaplus\paren{\state,\action}.}
\end{equation}
The approximation of the value function and gradient introduces bias 
and variance 
terms in the online learning regret bound ${\textrm{Regret}^{\mathcal{L}}_N}$. We propose a general theorem to lower bound the performance: %
\begin{proposition}[Adapted from \citet{cheng2020policy}]\label{prop:max_plus_performance_lowerbound} 
    Define $\Delta_N$, $\epsilon_{\ENum}\paren{\Policyclass}$, $\fplusAtM$, and ${\textrm{Regret}^{\mathcal{L}}_N}$ as above, where $\fplusAtM := \max_{k\in\bracket{\lvert \OracleSet \cup \{\policy_m\}\rvert}} \VFunc^{k}\paren{\state}$ for $m \leq N$, and ${\textrm{Regret}^{\mathcal{L}}_N}$ corresponds to the regret of a first-order online learning algorithm based on \eqnref{eq:max_plus_gradient}.  
    It holds that
    \begin{align*}
        \expct{\max_{n\in\bracket{\ENum}} \VFunc^{\policy_n} \paren{\stateDist_0}} \geq  \expctover{\state\sim\stateDist_0}{
        \fplusAtM\paren{\state}
        } + \expct{\Delta_N - \epsilon_{\ENum}\paren{\Policyclass} - 
         {\mathrm{Regret}^{\mathcal{L}}_N}},
    \end{align*}
    where the expectation is over the randomness in feedback and the online algorithm.
\end{proposition}

\begin{remark}\label{rem:max_plus_lower_bound}
    W.l.o.g. 
    we assume $\fplus_0(\state) := \argmax_{k\in\bracket{\ONum}} \; \VFunc^{k}\paren{\state}$, \fix{which corresponds to the baseline function considered in MAMBA.} Note that $\fplusAtM(\state)$ admits a weaker baseline value than $\fplus_n(\state)$ for $m<n$,  %
    but \textit{no weaker than} the max value of any oracle,  $\fplus_0(\state)$.
    \fix{Therefore, as the learner improves $\fplusAtM(\state)$, 
    \maxplusagg will have an improved lower bound over \citet{cheng2020policy}}. Consider a scenario where $m=o(N)$. In round $m$, %
    we instantiate $\fplusAtM$  %
    and perform 1-step advantage improvement over $\fplusAtM$.
    Since $\fplusAtM(\state) > \fplus_0(\state)$ when $\VFunc^{\policy_{m}(\state)} > \fplus_0(\state), \state\sim\stateDist^{\policy_n}$, %
    we can view \maxplusagg as adding improved learner policies into $\OracleSet$ at the end of each round and perform 1-step improvement over 
     $\fplus$ on the \emph{expending} oracle set. {As $\expctover{\state\sim\stateDist_0}{\fplusAtM\paren{\state}}$ improves}, it will lead to the improvement over the original bound in \propref{prop:max_plus_performance_lowerbound}.
\end{remark}

\vspace{-1mm}
\section{
Robust Policy Improvement 
 {via}
Actively 
Blending RL and IL}
\vspace{-1mm}

In this section, we present \algname, 
an algorithm for robust policy improvement {that builds upon the \maxplus-aggregation policy}. \algname consists of two main components: Robust Active Policy Selection~(RAPS) and Robust Policy Gradient~(RPG) that enable the algorithm to combine the advantages of reinforcement and imitation learning. %

\begin{algorithm}[!t]
    \caption{Robust Policy Improvement (\algname)} \label{alg:rpi}
    \begin{algorithmic}[1] 
        \Require Learner policy $\policy_{1}$, oracle set $\Policies=\curlybracket{\policy^{k}}_{k\in \bracket{\ONum}}$, function approximators $\{\hat{\VFunc}^k\}_{k\in\bracket{\ONum}}$, $\hat{\VFunc}_n$.
        \Ensure {The best policy among \curlybracket{\policy_1,...,\policy_{\ENum}}.}
        \For{$n=1, \ldots, N-1$} 
        \State Construct an extended oracle set $\ExtendOracleSet = \bracket{\policy^1, \ldots ,\policy^k, \policyN}_{k\in \bracket{|\Policies|}}$.
        \State Sample $t_e\in \bracket{\horizon-1}$ uniformly random.
        \State Roll-in $\policyN$ up to $t_e$, select $k_{\star}$ (\eqnref{eq:kstar}), %
        and roll out $\policy^{k_{\star}}$ to collect \fix{the remaining} data $\mathcal{D}^k$. \label{lin:riro}
        \State Update $\hat{V}^{k_{\star}}$ using $\mathcal{D}^k$.
        \State Roll-in $\policyN$ for full $\horizon$-horizon to collect data $\mathcal{D}_n'$.
        \State Update $\color{blue}\hat{V}_n$ using $\mathcal{D}_n'$.
        \State Compute advantage $\hat{\AFunc}^{ {\textrm{GAE}+}}$~
        (\eqnref{eq:A_Plus_GAE})
        and gradient estimate $\hat{\gradient}_n$~(\eqnref{eq:rpi_gradient}) using $\mathcal{D}_n'$.\label{lin:ac:begin}
        \State Update $\policyN$ to $\color{blue}\policy_{n+1}$ by giving $\hat{g}_n$  to a first-order online learning algorithm.\label{lin:ac:end}
        \EndFor 
    \end{algorithmic}
\end{algorithm}

\vspace{-1mm}
\subsection{Robust Active Policy Selection}
\vspace{-1mm}

To improve the sample efficiency in learning from multiple oracles and lower the bias %
in $\textrm{Regret}^{\mathcal{L}}_\ENum$ in \propref{prop:max_plus_performance_lowerbound}
caused by the 
{approximator} of the {\maxplus~baseline} function $\estfplus$, we propose a \textit{robust active policy selection} strategy. We employ an ensemble of prediction models to estimate the value function for a policy~\fix{\citep{liu2023active}}, where we estimate both the mean $\hat{\VFunc}_{\mu}^k\paren{\state}$ and the uncertainty $\sigma_k\paren{\state}$ for a particular state $\state$. We generate %
{a few}
independent value prediction networks that are initialized randomly, and then train them using random samples from the trajectory buffer of the corresponding oracle $\pi^k$. %

\looseness -1 In the single oracle case, the motivation of rolling in a learner policy and rolling out an oracle policy (referred to RIRO) in prior work (e.g., DAgger, AggrevateD) is to address the distribution shift.
In our work, in addition to addressing distribution shift, we aim to improve the value function estimator $\hat{\VFunc}$ of the most promising oracle on the switch state $\state$ to reduce the bias term of $\estfplus$.
Moreover, we seek to reduce the roll-out cost associated with querying oracles, particularly when the learner exhibits a higher expected value for the switching state. In such cases, we roll-out the learner to collect additional data to enhance its policy.
We achieve this goal  
by comparing the UCB of oracle policies' value function and LCB of learner policy to improve the estimation of $\estfplus$. We design the strategy as follows:

Let $\overline{{\hat{\VFunc}^k}}(\state) = \hat{\VFunc}_{\mu}^k(\state) + \sigma_k(\state), \underline{{\hat{\VFunc}^k}}(\state) = \hat{\VFunc}_{\mu}^k(\state) - \sigma_k(\state)$ be the UCB and LCB of policy $k$'s value function for state $\state$, respectively. We obtain the best oracle ${\policy^{k_{\star}}}$ for state $\state$ as follows:
\begin{equation}\label{eq:kstar}
 {k_{\star}} = \argmax_{k\in\bracket{|\ExtendOracleSet|}}\curlybracket{\overline{{\hat{\VFunc}^1}}\paren{\state},\overline{{\hat{\VFunc}^2}}\paren{\state},..., \overline{{\hat{\VFunc}^{\ONum}}}\paren{\state}, \underline{{\hat{\VFunc}^{\ONum+1}}}\paren{\state}},   
\end{equation}
where $\underline{\hat{\VFunc}^{\ONum+1}}$ is the confidence-aware value function approximator for the learner's policy, while $[\overline{{\hat{\VFunc}^k}}]_{k\in\bracket{\ONum}}$ represents the value function approximators associated with oracle policies.

\begin{remark}
The insight behind using a confidence-aware policy selection strategy in RAPS is to %
improve the estimate of the value function of the most promising oracle at a given state. This necessitates accounting for estimation uncertainties, which leads to the adoption of a UCB-based approach to identify the optimal oracle. Using LCB for the learner encourages \fix{oracle-guided} exploration %
unless we are certain that the learner surpasses all oracles for the given state. We empirically evaluate this in Section~\ref{exp:RAPS:confidence}. %
\end{remark}

\begin{remark}
{MAPS~\citep{liu2023active} introduced an active policy selection strategy by selecting the best oracle to roll out and improve the value function approximation on %
state $\state_{t_e}$ according to $\fplus_0$. %
In this work, we {empirically} improve such %
strategy by utilizing the learner policy in $\ExtendOracleSet$. %
}
\end{remark}

\subsection{Robust Policy Gradient}

We now propose robust policy {gradient}
based on a novel advantage function, denoted by $\aplusGAE$ 
and a novel \maxplus~actor-critic framework.

$\aplusGAE$~\textbf{advantage function.} The policy gradient methods maximize the expected total reward by repeatedly estimating the gradient $\gradient:= \triangledown_{\theta}\mathbb{E}[\sum_{t=0}^{{\horizon-1}}\reward_t]$. %
The policy gradient has the form 
${\gradient}={\mathbb{E}}_{t}[\triangledown_{\theta}{\log{\policy_{\theta}\paren{\action_t|\state_t}\hat{\AFunc}_t }}]$~\fix{\citep{sutton1999policy, greensmith2004variance, schulman2015high,schulman2017proximal}}, where $\policy_{\theta}$ is a stochastic learner policy and $\hat{\AFunc}_t$ is an estimator of the advantage function at timestep $t$ and $\mathbb{E}[\cdot]$ indicates the empirical average over a finite batch of samples, for an algorithm that alternates between sampling and optimization. 
$\AFunc_t$ measures whether the action is better or worse than the current policy.
Hence, the gradient term $\triangledown_{\theta}{\log{\policy_{\theta}\paren{\action_t|\state_t}\hat{\AFunc}_t }}$ points in the direction of increased $\policy_{\theta}\paren{\action_t|\state_t}$ if and only if $\hat{\AFunc}_t=\hat{\AFunc}\paren{\state_t,\action_t}>0$. %
For $\hat{\AFunc}$, we propose a novel advantage function $\aplusGAE$ based on general advantage estimation~\citep{schulman2015high}, the  {\maxplus~baseline $\fplus$} 
and the $\aplus$ advantage function~\eqref{eq:A_plus_advantage_function}. 
\begin{equation}\label{eq:A_Plus_GAE}
\hat{\AFunc}^{ {\textrm{GAE}\paren{\gamma,\lambda}\texttt{+}}}_t=\hat{\delta}_t+\paren{\gamma\lambda} {\hat{\delta}_{t+1}}+...+\paren{\lambda\gamma}^{ {T}-t+1}\hat{\delta}_{ {T}-1},   ~\textrm{where}~  {\hat{\delta}}_t=\reward_t+\gamma\estfplus\paren{\state_{t+1}} - \estfplus\paren{\state_t},
\end{equation}%
where $T \ll H$, %
and $\gamma$ and $\lambda$ are the predefined parameters that control the bias-variance tradeoff. 
In this work, we use $\lambda=0.9$ and $\gamma=1$, and thus simplify $\hat{\AFunc}^{\textrm{GAE}\paren{\gamma,\lambda}\texttt{+}}_t$ as $\estaplusGAEt$. %

We propose a variant of the 
 {\maxplus~baseline $\fplus$}
that includes a confidence threshold $\threshold$ for an oracle's value estimate:
\fix{
\begin{equation}\label{eq:f_plus_confidence_aware}
   \!\!\!\! \estfplus\paren{\state}=
    \begin{cases}
            \hat{\VFunc}^{\fix{\policy_n}}_{\mu}\paren{\state}, \quad~\textrm{if}~\sigma_{k}\paren{\state}>\threshold, \textrm{where } k=\argmax_{k\in\bracket{|\ExtendOracleSet|}} 
            \hat{\VFunc}^k_{\mu}\paren{\state}.\\
        \max_{k\in{\bracket{|\ExtendOracleSet|}}}\hat{\VFunc}^k_{\mu}\paren{\state}, \quad \text{otherwise}.
    \end{cases}
\end{equation}
}
\begin{remark}
We use a threshold to control the reliability of taking the advice of an oracle, where a lower value indicates greater confidence. %
In our experiments, we use $\threshold=0.5$, which we have found to exhibit robust behavior (\appref{app:tuning_gamma}).
\end{remark}

\begin{figure}[t!]
    \centering
    \includegraphics[width=.95\textwidth]{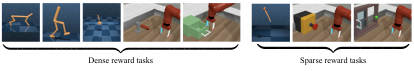}
    \vspace{-3mm}
    \caption{%
      We consider 8 tasks from DeepMind Control Suite and Meta-World. %
     Extended results on different variants of these tasks are provided in Appendices~\ref{sec:metaworld-dense} \& \ref{sec:metaworld-sparse}. %
}\label{fig:exp:env}
\end{figure}

Finally, we have {the $n$-th  {round} online loss as
\begin{equation}\label{eq:rpi_loss}
    \hat{\loss}_{n}\paren{\policy_n}:=-\horizon\mathbb{E}_{\state\sim\stateDist^{\policy_n}} { {{\mathbb{E}}_{\action\sim {\policy|\state}}}}
    \bracket{ {\estaplusGAE\paren{\state,\action}}}|_{\policy=\policy_n},
\end{equation}
}
and gradient estimator as
\begin{equation}\label{eq:rpi_gradient}
    {\hat{\gradient}_n} =\nabla \hat{\loss}_{n}\paren{\policy_n} = {-\horizon}{\mathbb{E}}_{\state\sim\stateDist^{\policy_n}}{\mathbb{E}}_{\action\sim {\policy|\state}} \bracket{\nabla{\log{\policy \paren{\action \mid \state}  {\estaplusGAEt\paren{\state,\action}} }}}|_{\policy=\policy_n}.
\end{equation}
\paragraph{\Maxplus~actor-critic.}
We note that the RPG component (\algoref{alg:rpi}, lines~\ref{lin:ac:begin}--\ref{lin:ac:end}) can be viewed as a variant of the \textit{actor-critic} framework, with the actor sampling trajectories that are then evaluated by the \maxplus critic based on the
$ {\aplusGAE}$
advantage function \eqref{eq:A_Plus_GAE}. 
The policy gradient in Eqn.~\ref{eq:rpi_gradient} enables the learner policy $\pi_n$ to learn from high-performing oracles and to {improve} %
its own value function $\hat{V}^k$ for the states in which the oracles perform poorly.

\begin{remark}\label{thm:rpi_performance_lower_bound}
When $\gamma=1$, Eqn.~\ref{eq:A_Plus_GAE} disregards the accuracy of $\estfplus$,
but it has high variance due to the sum of the reward terms. When $\gamma=0$, it introduces bias, but has much lower variance. Moreover,
when $\lambda=0$ and $\gamma=1$, the loss \eqref{eq:rpi_loss}
of \algname reduced to the loss \eqref{eq:max_plus_loss} under \maxplusagg \eqref{eq:max_plus_aggregation}, and the performance bound for the $\maxplusagg$ policy and \algname will be equal. Thus, performing no-regret online learning with regards to Eqn.~\ref{eq:rpi_loss} has the guarantee in
\propref{prop:max_plus_performance_lowerbound}
and \remarkref{rem:max_plus_lower_bound}.
However, when $\lambda>0$, \algname will 
 {optimize} the 
multi-steps advantage
over $\fplus$ in  Eqn.~\ref{eq:rpi_loss}, while the \maxplusagg policy \polmaxplusagg only 
 {optimizes} the 
one-step advantage
over $\fplus$. Thus, \algname will have a smaller $\epsilon_{\ENum}\paren{\Policyclass}$ term than \maxplusagg, 
which improves the performance lower bound in \propref{prop:max_plus_performance_lowerbound}.
\end{remark}

\textbf{Imitation, Blending and Reinforcement.} 
Instances of $\estfplus$ in Eqn.~\ref{eq:A_Plus_GAE} may involve a combination of oracles' and learner's value functions. In a case that this does not involve the learner's value function---this is likely in the early stage of training since the learner's performance is poor---\algname performs imitation learning on the oracle policies.
Once the learner policy improves and $\estfplus$ becomes identical to the learner's value function, \algname becomes equivalent to the vanilla actor-critic that performs self-improvement.
When it is a combination of the two, \algname learns from a blending of the learner and oracles.

\vspace{-8mm}
\section{Experiments}
\vspace{-1mm}

\begin{figure*}[t!]
    \rotatebox[origin=l]{90}{\quad\quad\quad \scriptsize DMC Suite}
    \begin{subfigure}{.24\textwidth}
        \centering
        \includegraphics[%
        width=3.4cm,  clip={0,0,0,0}]{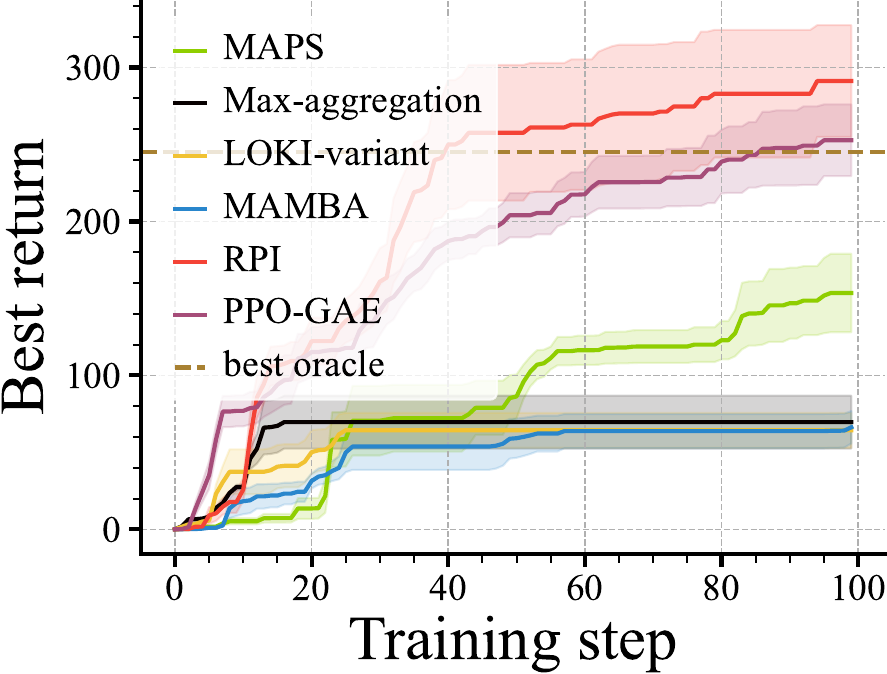}
        \caption{Cheetah (dense)}\label{fig:results:cheetah}
    \end{subfigure}\hfil
    \begin{subfigure}{.24\textwidth}
        \centering
        \includegraphics[%
        width=3.4cm,  clip={0,0,0,0}]{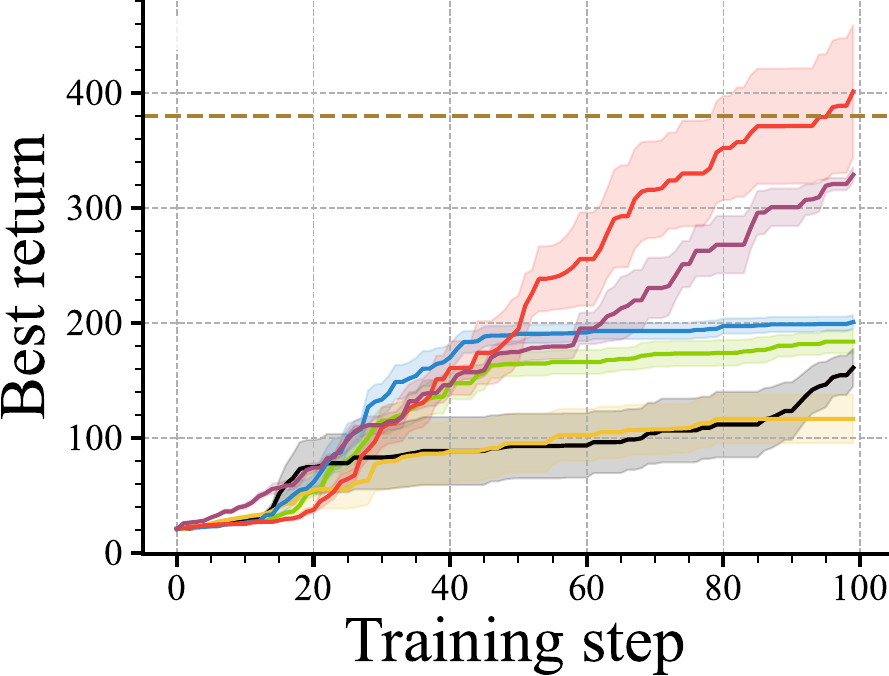}
        \caption{Walker-walk (dense)}\label{fig:results:walker}
    \end{subfigure}\hfil
    \begin{subfigure}{.24\textwidth}
        \centering
        \includegraphics[%
        width=3.4cm,  clip={0,0,0,0}]{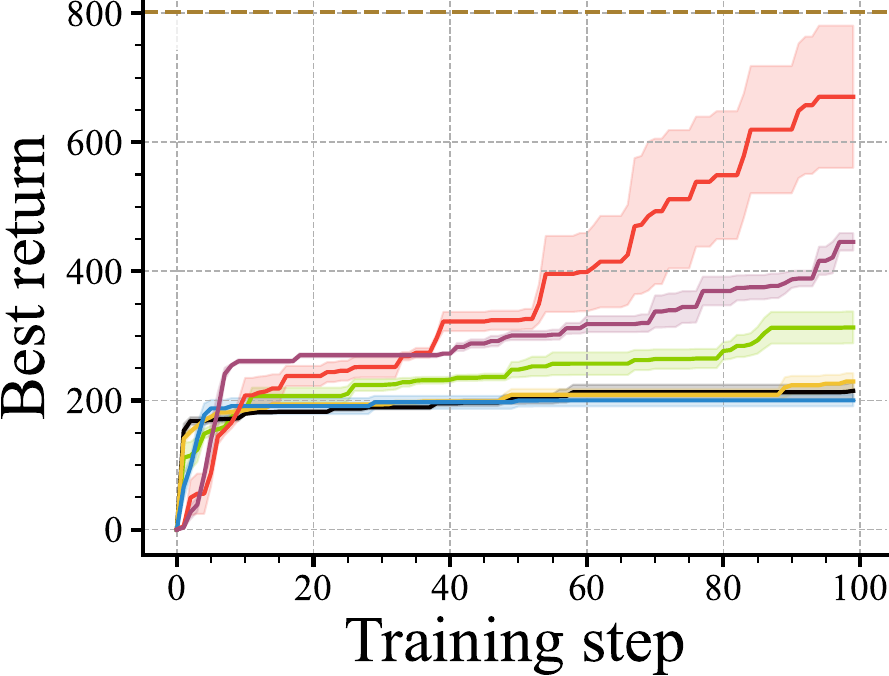}
        \caption{Cartpole (dense)}\label{fig:results:cartpole}
    \end{subfigure}\hfil
    \begin{subfigure}{.24\textwidth}
        \centering
        \includegraphics[%
        width=3.4cm,  clip={0,0,0,0}]{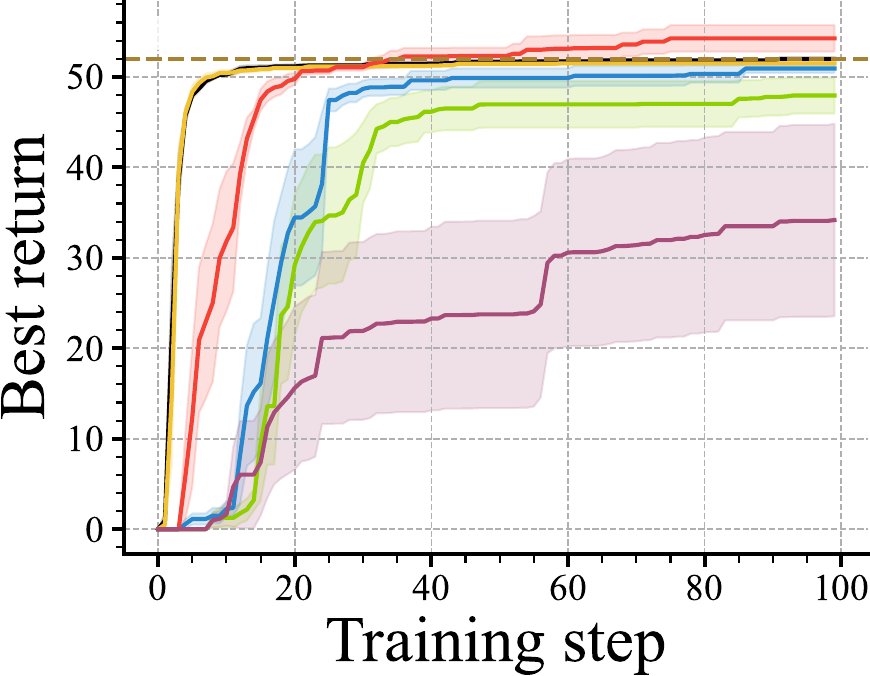}
        \caption{Pendulum (sparse)}\label{fig:results:pendulum}
    \end{subfigure}\hfil

    \rotatebox[origin=l]{90}{\quad\quad\quad \scriptsize Meta-World}
    \begin{subfigure}{.24\textwidth}
        \centering
        \includegraphics[%
        width=3.4cm,  clip={0,0,0,0}]{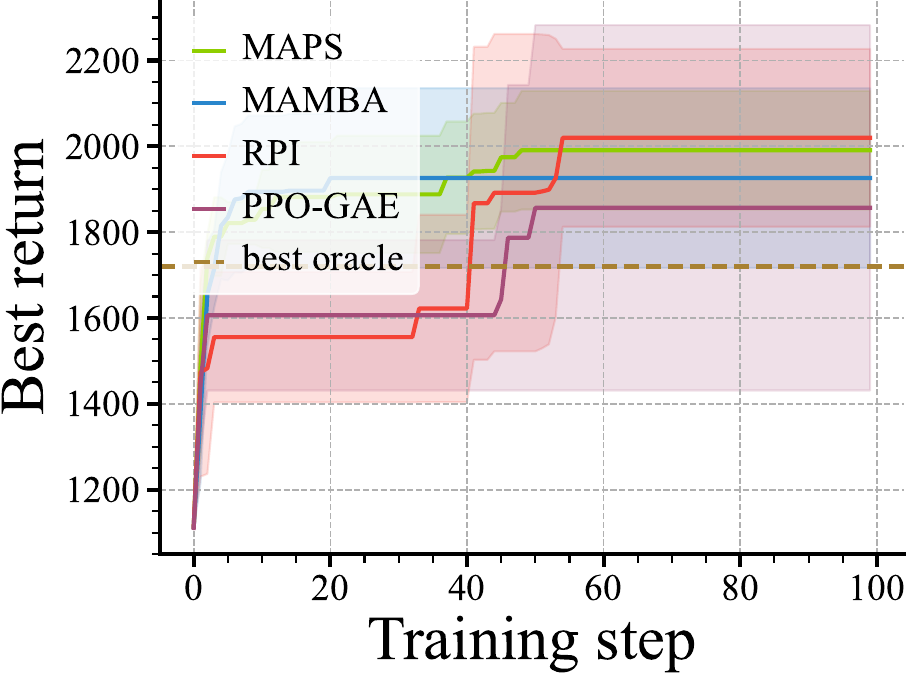}
        \caption{Faucet-open~(dense)}\label{fig:results:faucet-open:main}
    \end{subfigure}\hfil
    \begin{subfigure}{.24\textwidth}
        \centering
        \includegraphics[%
        width=3.4cm,  clip={0,0,0,0}]{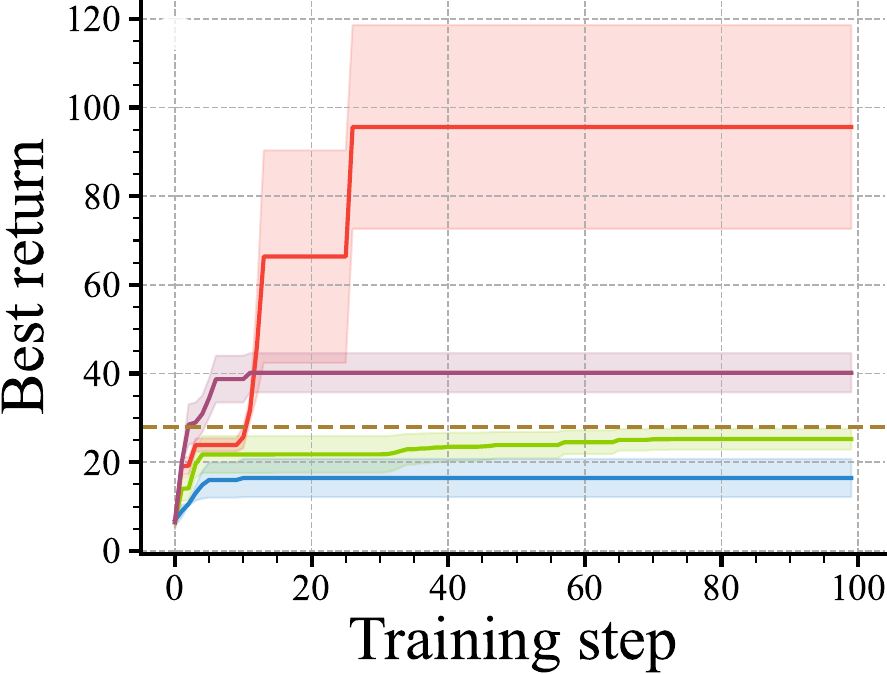}
        \caption{Drawer-close~(dense)}\label{fig:results:drawer-close:main}
    \end{subfigure}
    \begin{subfigure}{.24\textwidth}
        \centering
        \includegraphics[%
        width=3.4cm,  clip={0,0,0,0}]{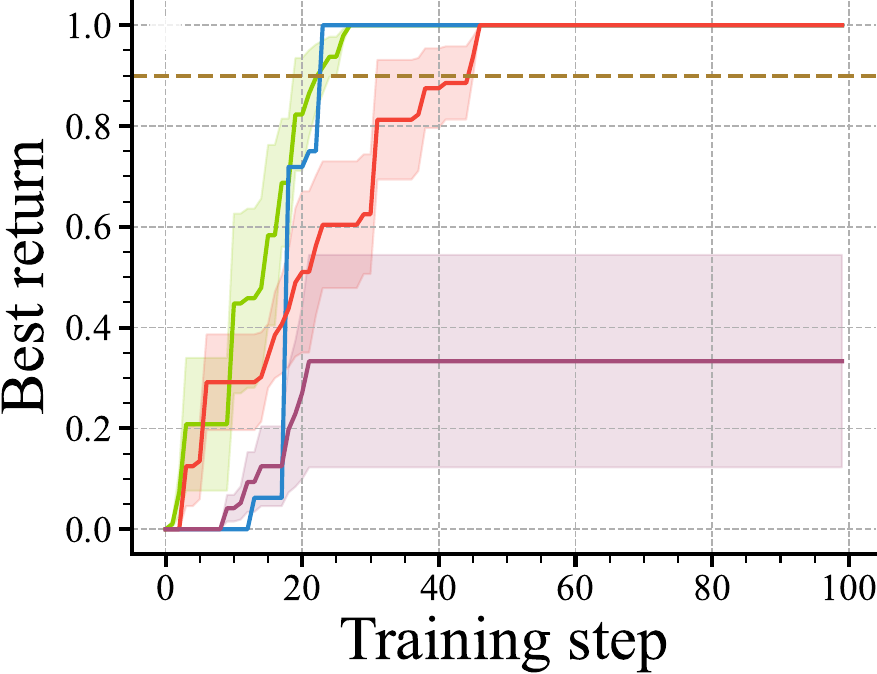}
        \caption{Button-press~(sparse)}\label{fig:results:button-press:main}
    \end{subfigure}\hfil
    \begin{subfigure}{.24\textwidth}
        \centering
        \includegraphics[%
        width=3.4cm,  clip={0,0,0,0}]{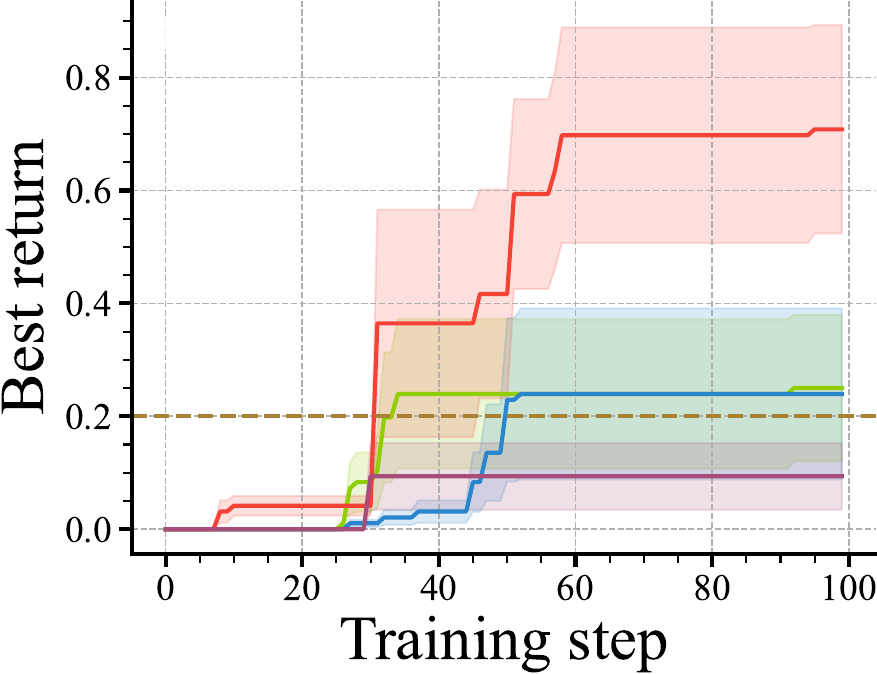}
        \caption{Window-close~(sparse)}\label{fig:results:window-close:main}
    \end{subfigure}\hfil
    \caption{\textbf{Main results.} A comparison between \algname with five baselines 
    and the best oracle (horizontal line)
    on Cheetah, Cartpole, Pendulum, and Walker-walk from DMC;
    and Window-close, Button-press, Faucet-open, and Drawer-close from Meta-World
    in terms of best-return %
    with \reminder{three diversified oracles}. 
    The shaded area represents the standard error over five 
    trials. \algname %
    scores the best in all benchmarks.
    }\label{fig:exp:multi_experts}
    \vspace{-3mm}
\end{figure*}

\vspace{-2mm}
\paragraph{Environments.}
We evaluate our method on eight continuous state and action space domains: Cheetah-run, CartPole-swingup, Pendulum-swingup, and Walker-walk from the DeepMind Control Suite~\citep{tassa2018deepmind}; and Window-close, Faucet-open, Drawer-close and Button-press from Meta-World~\citep{yu2020meta}. In addition, we conduct experiments on a modified sparse reward Meta-World environment, which is considered to be a more challenging task. \fix{We set ensemble size of value functions 
to five.}
Appendix~\ref{app:experiment} provides further details.

\vspace{-3mm}
\paragraph{Oracles.} We implement our oracles as policies trained using PPO~\citep{schulman2017proximal} with generalized advantage estimate (GAE)~\citep{schulman2015high} and SAC~\citep{haarnoja2018soft}. %
We save the policy weights at different points during training to achieve
oracles that perform differently in different states. \fix{Each environment is provided with three diversified oracles.}

\vspace{-3mm}
\paragraph{Baselines.}
We compare \algname with five baselines:
(1) PPO with GAE as a pure RL baseline;
(2) Max-Aggregation~\citep{cheng2020policy} as a pure IL baseline (a multiple-oracle variant of AggreVaTe(D));
(3) a variant of LOKI
adapted to the multiple-oracle setting that initially performs pure IL and then pure RL; %
(4) MAMBA;
(5) MAPS (the current state-of-the-art method)\footnote{Our experimental setup  {including the oracle set} differs from that of MAPS. In this work, the learner for all baselines has access to approximately the same number of transitions and the learner does not have access to the oracle's trajectory.  {We reproduce the baseline performance %
for the MAPS' setting in \appref{app:baseline_alignment}}.}; %
and also the best oracle in the oracle set as a reference. 
We matched the number of environment interactions across algorithms\footnote{\fix{PPO has a slightly smaller number of interactions due to the lack of oracles' value function pre-training.}}.
Appendix~\ref{app:experiment} provides further details.

\vspace{-2mm}
\subsection{Main Results}
\vspace{-1mm}
Figure~\ref{fig:exp:multi_experts} visualizes the performance of \algname and the baselines. %
The results show that \algname surpasses the baselines on all domains, despite variations in the black-box oracle set. Notably, the RL-based PPO-GAE baseline outperforms the IL methods in the later stages of training  {in most of the dense reward environments,  %
while IL-based approaches perform better in the sparse reward domains
}.
Pendulum-swingup~(\figref{fig:results:pendulum})  {and window-close~(\figref{fig:results:window-close:main}) are} particularly difficult domains that involve non-trivial dynamics and sparse reward (i.e., the agent receives a reward of $1$ only when the pole is near vertical,  {the window is closed exactly}). 
Due to the sparse reward, the IL-based approaches are significantly more sample efficient than the RL-based approach, but their performance plateaus quickly.
 {\algname initially bootstraps from the oracles, and due to their suboptimality, it switches to self-improvement (i.e., learning from its own value function), resulting in better performance than both IL and RL methods.} %
These results demonstrate the robustness of \algname as it actively combines the advantages of IL and RL to adapt to various environment.

\vspace{-2mm}

\subsection{Ablation studies}

\begin{figure*}[t!]
    \begin{subfigure}{.24\textwidth}
        \centering
        \includegraphics[%
        width=3.4cm,  clip={0,0,0,0}]{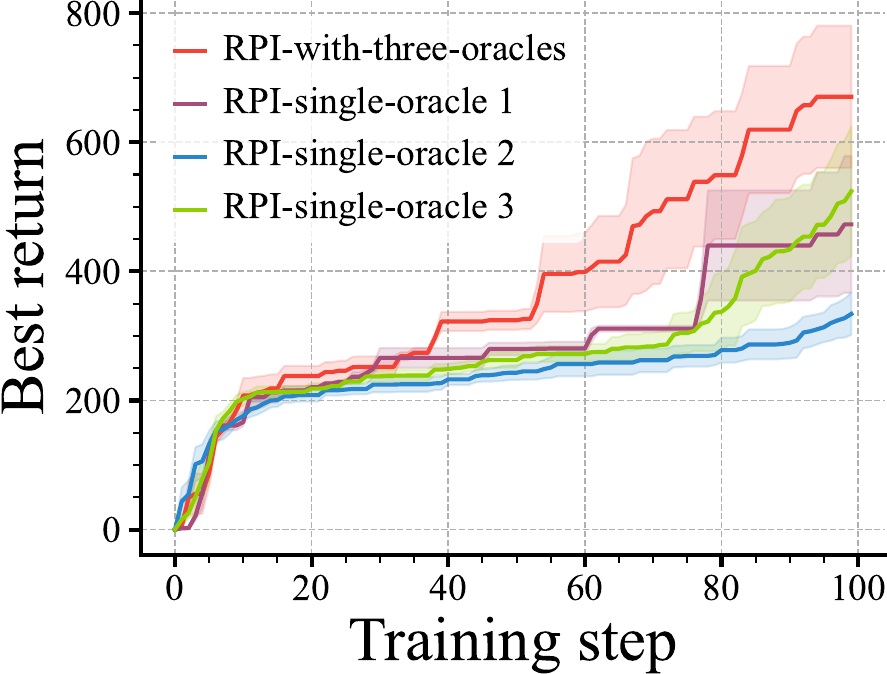}
        \caption{Learning state-wise expertise.}\label{fig:exp:combine_expertise}
    \end{subfigure}\hfil
    \begin{subfigure}{.24\textwidth}
        \centering
        \includegraphics[%
        width=3.4cm,  clip={0,0,0,0}]{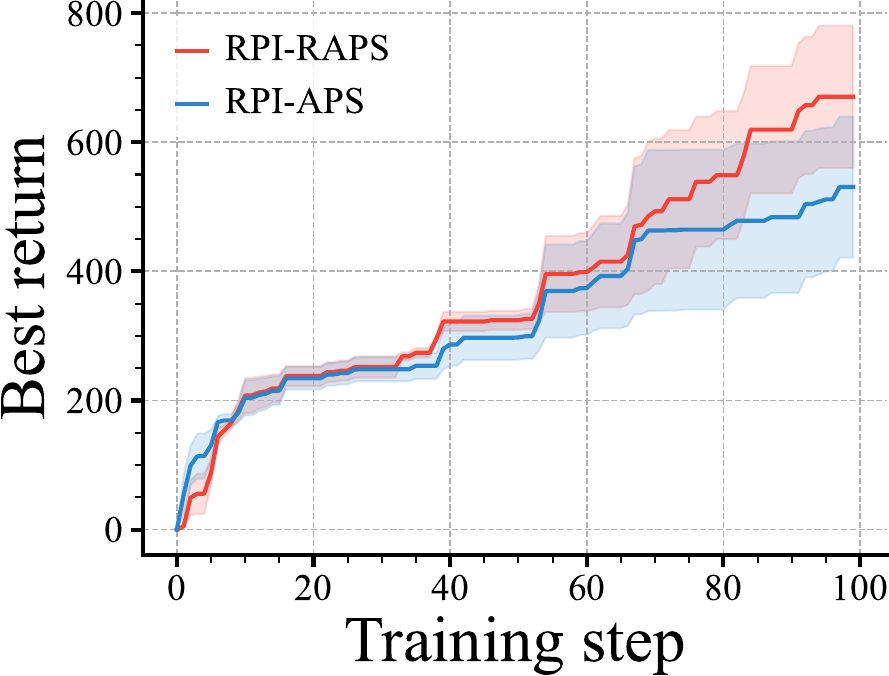}
        \caption{RAPS vs APS.}\label{fig:exp:sample}
    \end{subfigure}\hfil
    \begin{subfigure}{.24\textwidth}
        \centering
        \includegraphics[%
        width=3.4cm,  clip={0,0,0,0}]{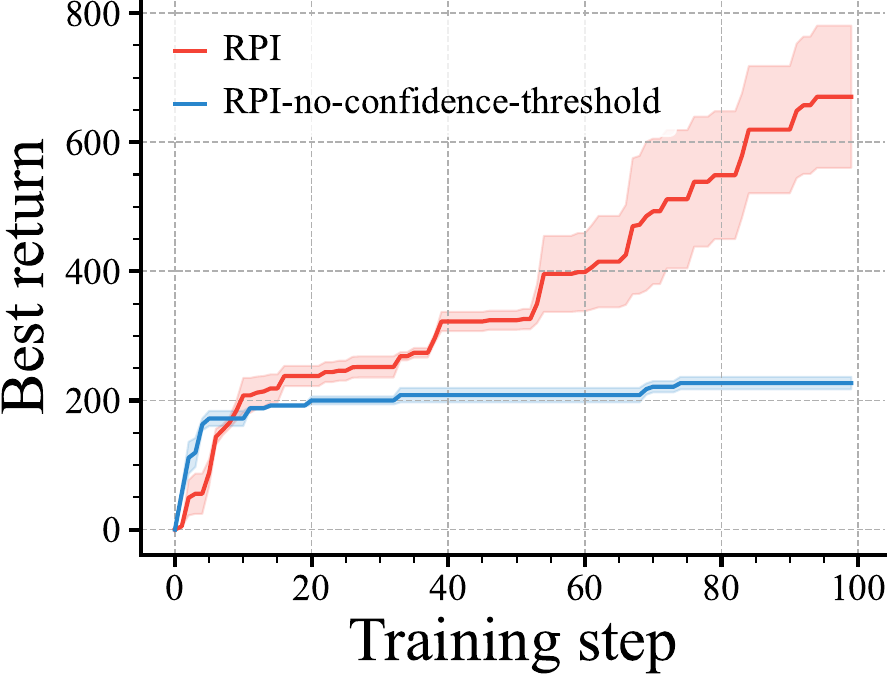}
        \caption{ {Confidence-aware RPG}}\label{fig:Confidence-awareness}
    \end{subfigure}\hfil
    \begin{subfigure}{.24\textwidth}
        \centering
        \includegraphics[%
        width=3.4cm,  clip={0,0,0,0}]{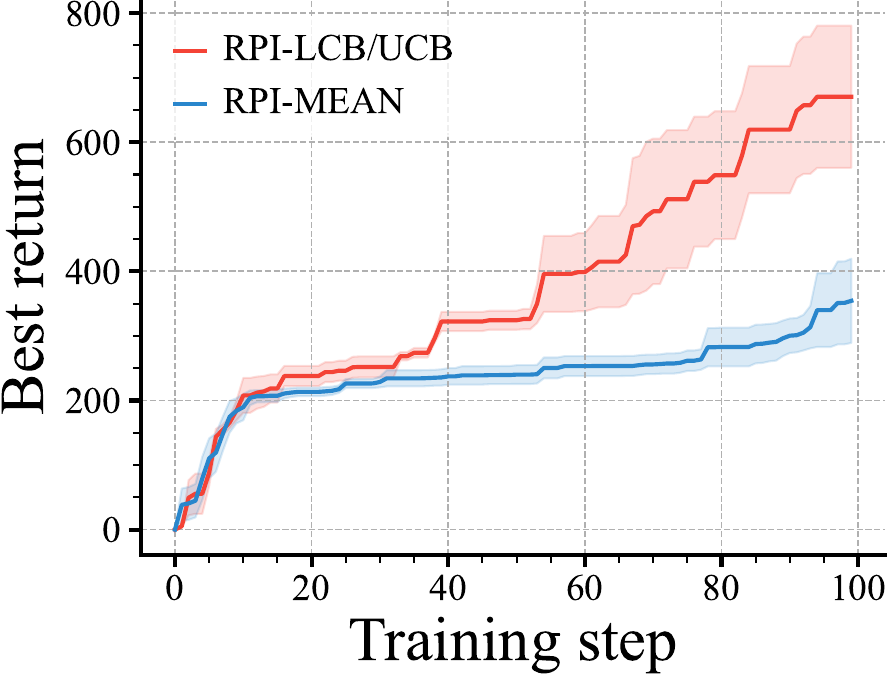}
        \caption{ {Confidence-aware RAPS}}\label{fig:exp:raps-confidence}
    \end{subfigure}\hfil
    \caption{Results of ablation studies on the Cartpole environment.}\label{fig:exp:ablation}
\end{figure*}

\paragraph{ {Learning}
state-wise oracle expertise.} In \figref{fig:exp:combine_expertise}, we examine the ability of \algname to aggregate the expertise of multiple oracles %
on Cartpole. 
We created three diversified oracles and find that RPI achieves a return of $645$ when it is able to query all three oracles, while the  {best} return falls below $600$ when given access to only a single oracle. 
This result demonstrates \algname's utilization of the oracles' state-wise expertise, as it achieves better performance when given access to more oracles. %

\vspace{-2mm}
\paragraph{Ablation on robust active policy selection.
} In order to understand the effectiveness of RPI's robust active policy selection strategy (RAPS), we compare it to active policy selection (APS)~\citep{liu2023active} (without the  {learner} in RIRO~(\algoref{alg:rpi}, line \ref{lin:riro})) on Cartpole.
{\figref{fig:exp:sample} shows that RAPS has the advantage of selecting the learner policy to roll out in states for which it outperforms the oracles, resulting in self-improvement. This leads to RAPS outperforming the APS-based approach.}
\vspace{-3mm}
\paragraph{Confidence-awareness in RPI.}\label{exp:RAPS:confidence}
\emph{(1) RPG}: We first perform an ablation on Cartpole to investigate the benefits of using a confidence threshold on an oracle's value estimate for RPG (Eqn.~\ref{eq:f_plus_confidence_aware}). We see in \figref{fig:Confidence-awareness} that the confidence threshold enables RPG to benefit from both state-wise imitation learning from oracles with high confidence and the execution of reinforcement learning when oracles exhibit high uncertainty. Without the threshold, RPG is more vulnerable to the quality of oracle set.
\emph{(2) RAPS}: We then consider the benefits of reasoning over uncertainty to the policy selection strategy, comparing uncertainty-aware RPI-LCB/UCB (Eqn.~\ref{eq:kstar}) to RPI-MEAN, which does not consider uncertainty. \figref{fig:exp:raps-confidence} demonstrates the benefits of using LCB/UCB for policy selection. Addition results in \appref{app:lcb-ucb-raps} reveal that RPI-LCB/UCB outperforms RPI-MEAN across all benchmarks by an $\textit{overall}$ margin of 40\%, supporting the advantage of incorporating confidence to policy selection.

\vspace{-2mm}
\begin{wrapfigure}[9]{r}{.26\textwidth}
    \vspace{-10pt}
    \centering
    \includegraphics[width=\linewidth]{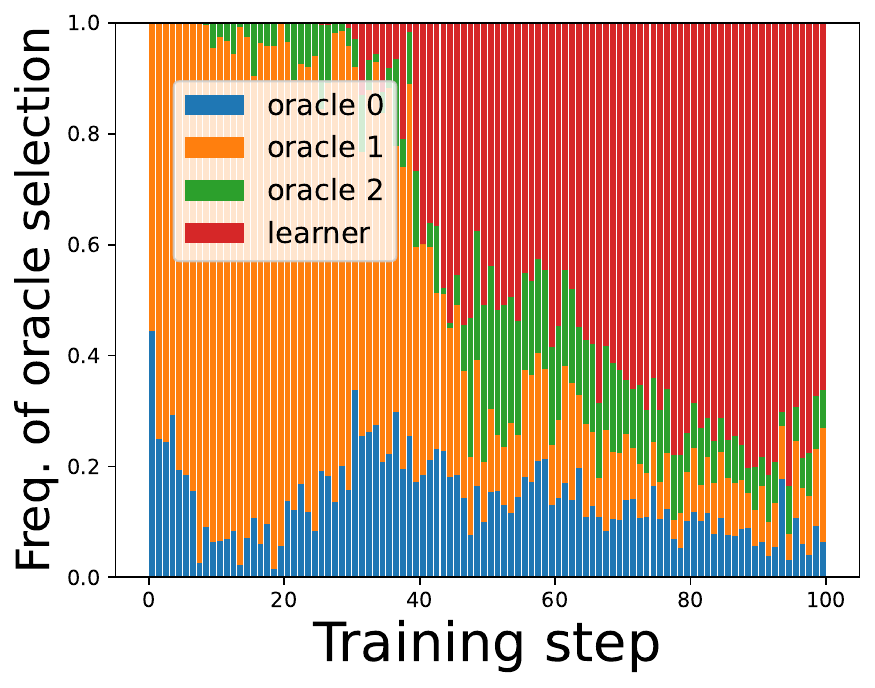}
    \vspace{-7mm}
    \caption{IL and RL.}\label{fig:exp:imitation_reinforcement}
\end{wrapfigure}
\paragraph{Visualizing {active} IL and RL.} Figure~\ref{fig:exp:imitation_reinforcement} 
visualizes the {active} state-wise imitation and reinforcement process employed by \algname in the gradient estimator on Pendulum. The figure includes three oracle policies (in \textcolor{myblue}{blue}, \textcolor{myorange}{orange}, and \textcolor{mygreen}{green}) and the learner's policy (in \textcolor{myred}{red}). Each oracle exhibits different expertise at different stages. In the beginning, \algname only imitates the oracles, which initially have greater state-wise expertise than the learner. As the learner improves, the frequency with which \algname samples the leaner policy increases, corresponding to self-improvement. As training continues, the expertise of the learner increasingly exceeds that of the oracles, resulting in \algname choosing self-improvement more often than imitating the oracles.

\vspace{-2mm}
\section{Conclusion}
\vspace{-2mm}
We present {\maxplus}, a robust framework for IL and RL in the presence of a set of black-box oracles. 
{Within this framework,}
we introduce \algname, a policy gradient algorithm comprised of two novel components: a robust active policy selection strategy (RAPS) that enhances sample efficiency and a robust policy gradient (RPG) for policy improvement.
We provide %
a rigorous theoretical analysis of \algname, demonstrating its superior performance compared to the current state-of-the-art. %
Moreover, empirical evaluations on a diverse set of tasks demonstrate that \algname consistently outperforms all IL and RL baselines,
even in scenarios with limited oracle information (favoring RL) or sparse rewards (favoring IL). \algname effectively adapts to the nature of the domain and the quality of the oracles by actively interleaving %
{IL and RL}. %
Our work introduces new avenues for robust imitation and reinforcement learning and encourages future research on addressing more challenging tasks in robust settings, such as handling missing state or oracle information. %

\section*{Acknowledgements}
We thank Ching-An Cheng for constructive suggestions. This work is supported in part by the RadBio-AI project (DE-AC02-06CH11357), U.S. Department of Energy Office of Science, Office of Biological and Environment Research, the Improve project under contract (75N91019F00134, 75N91019D00024, 89233218CNA000001, DE-AC02-06-CH11357, DE-AC52-07NA27344, DE-AC05-00OR22725), the Exascale Computing Project (17-SC-20-SC), a collaborative effort of the U.S.\ Department of Energy Office of Science and the National Nuclear Security Administration, the AI-Assisted Hybrid Renewable Energy, Nutrient, and Water Recovery project (DOE DE-EE0009505), and NSF HDR TRIPODS (2216899).

\bibliographystyle{plainnat}
\bibliography{reference}

\clearpage

\appendix

\ifSubfilesClassLoaded{%
\title{Blending Imitation and Reinforcement Learning\\ for Robust Policy Improvement\\[10pt]%
{\Large Supplementary Material}}

\maketitle
}

\section{Selective Comparison against Related Works}\label{App:related}

\begin{table*}[ht!]
    \caption{A qualitative comparison of related algorithms. The publication years are included in parentheses for reference. Algorithms designed to fit a particular criterion are marked by ``$\checkmark$''; criteria that are not explicitly considered in the algorithm design are marked by ``$\times$''. }
    \label{table:alg_characteristics}   
    \centering
    {\scriptsize
    \scalebox{0.85}{
    \begin{tabular}{l l c c c c c c c c}
        \toprule
        \textbf{Algorithm}
        & \textbf{Criterion}
        & \textbf{Online}
        & \textbf{Stateful}
        & \textbf{Active}
        & \textbf{Interactive}
        & \textbf{\makecell[c]{Multiple\\ oracles}}
        & \textbf{\makecell[c]{Sample \\ efficiency\\ (in multiple\\oracles)}}
        & \textbf{Robust}
        \\
        \midrule
        {\makecell[l]{Behavioral Cloning \\{\citep{pomerleau1988alvinn}}}}
        &  IL 
        & {$\times$ }
        & {$\checkmark$}
        & {$\times$ }
        & {$\times$ }
        & {$\times$ }
        & {{--}}
        & {$\times$ }\\[10pt]
        {\makecell[l]{REINFORCE \\ \citep{williams1992simple}\\ \citep{sutton1999policy}}}
        &  RL 
        & {$\checkmark$}
        & {$\checkmark$}
        & {$\times$ }
        & {$\times$ }
        & {$\times$ }
        & {$\times$ }
        & {--}\\[15pt]
        {\makecell[l]{SMILe \\ \citep{ross2010efficient}}}
        &  IL 
        & {$\times$ }
        & {$\checkmark$}
        & {$\times$ } 
        & {$\times$ }
        & {$\times$ }
        & {{--}}
        & {$\times$ }\\[10pt]
        {\makecell[l]{DAgger \\ \citep{ross2011reduction}}}
        & IL
        & {$\checkmark$}
        & {$\checkmark$}
        & {$\times$ } 
        & {$\checkmark$}
        & {$\times$ }
        & {{--} }
        & {$\times$ }\\[10pt]
        {\makecell[l]{PPO with GAE \\ \citep{schulman2017proximal} \\ \citep{ schulman2015high}} }
        & RL
        & {$\checkmark$}
        & {$\checkmark$}
        & {$\times$ } 
        & {$\times$ }
        & {$\times$ }
        & {$\times$ }
        & {-- }\\[10pt]
        {\makecell[l]{AggreVateD \\ {\citep{sun2017deeply}} }}
        & IL
        & {$\checkmark$}
        & {$\checkmark$}
        & {$\times$ } 
        & {$\checkmark$}
        & {$\times$ }
        & {{--}  }
        & {$\times$ }\\[10pt]

        {\makecell[l]{\textrm{\fix{DQfD}}\\ \citep{hester2018deep}}}
        & \fix{Offline $\rightarrow$ online RL}
        & {$\checkmark$}
        & {$\checkmark$}
        & {{--} } 
        & {$\checkmark$}
        & {$\times$ }
        & {{--}  }
        & {{--}}\\[10pt]
        {\makecell[l]{\textrm{THOR }\\ \citep{sun2018truncated}}}
        & IL$+$RL
        & {$\checkmark$}
        & {$\checkmark$}
        & {$\times$ } 
        & {$\checkmark$}
        & {$\times$ }
        & {{--}  }
        & {$\times$}\\[10pt]
        \makecell[l]{\textrm{LOKI}\\ \citep{cheng2018fast}}
        & IL$+$RL
        & {$\checkmark$}
        & {$\checkmark$}
        & {$\times$ } 
        & {$\checkmark$}
        & {$\times$ }
        & {{--} }
        & {$\times$}\\[10pt]
        \makecell[l]{\textrm{ {SAC-X }}\\ \citep{riedmiller2018learning}}
        & RL
        & {$\checkmark$}
        & {$\checkmark$}
        & {$\times$ } 
        & {$\checkmark$}
        & {$\checkmark$ }
        & {$\times$ }
        & {$\times$}\\[10pt]
        {\makecell[l]{LEAQI\\ {\citep{brantley2020active}}  }}
        & IL
        & {$\checkmark$}
        & {$\checkmark$} 
        & {$\checkmark$}
        & {$\checkmark$}
        & {$\times$ }
        & {{--}}
        & {$\times$ }\\[10pt]
        {\makecell[l]{MAMBA\\ {\citep{cheng2020policy}}  }}
        &  IL$+$RL 
        & {$\checkmark$}
        & {$\checkmark$}
        & {$\times$ } 
        & {$\checkmark$}
        & {$\checkmark$}
        & {$\times$ }
        & {$\times$ }\\
        {\makecell[l]{A-OPS\\ {\citep{konyushova2021active}} }}
        & \makecell[l]{Policy Sel.}
        & {$\times$}
        & {$\times$}
        & {$\checkmark$}
        & {$\times$ }
        & {$\checkmark$}
        & {$\checkmark$}
        & {--}\\[10pt]
        {\makecell[l]{ILEED \\ {\citep{ilbelaiev}}}}
        & IL
        & {$\times$}
        & {$\checkmark$}
        & {$\times$ } 
        & {$\times$ }
        & {$\checkmark$ }
        & {$\times$ }
        & {$\times$ }\\[10pt]
        {\makecell[l]{ \fix{IQL}\\ {\citep{kostrikov2021offline}} }}
        & \makecell[l]{ \fix{Offline $\rightarrow$ online RL}}
        & {$\checkmark$}
        & {$\checkmark$}
        & {--}
        & {$\checkmark$}
        & {$\times$}
        & {--}
        & {--}\\[10pt]
        {\makecell[l]{CAMS\\ {\citep{liu2022cost}} }}
        & \makecell[l]{Model Sel.}
        & {$\checkmark$}
        & {$\times$}
        & {$\checkmark$}
        & {$\times$ }
        & {$\checkmark$}
        & {$\checkmark$}
        & {$\checkmark$ }\\
        {\makecell[l]{ {\fix{MoDem}}\\ \citep{hansen2022modem}}}
        & \fix{Offline $\rightarrow$ online RL}
        & {$\checkmark$}
        & {$\checkmark$}
        & {--}
        & {$\checkmark$}
        & {$\times$}
        & {--}
        & {--}\\[10pt]
        {\makecell[l]{ {\fix{Hybrid RL}}\\ \citep{song2022hybrid}}}
        & \fix{Online RL with offline data}
        & {$\checkmark$}
        & {$\checkmark$}
        & {{--}}
        & {$\checkmark$}
        & {$\times$}
        & {--}
        & {--}\\[10pt]
        {\makecell[l]{ \fix{PEX}\\ \citep{zhang2023policy}}}
        & \fix{Offline $\rightarrow$ online RL}
        & {$\checkmark$}
        & {$\checkmark$}
        & {--}
        & {$\checkmark$}
        & {$\times$}
        & {--}
        & {--}\\
        {\makecell[l]{ {TGRL}\\ \citep{shenfeld2023tgrl}}}
        & IL$+$RL
        & {$\checkmark$}
        & {$\checkmark$}
        & {$\times$}
        & {$\checkmark$}
        & {$\times$}
        & {--}
        & {$\times$ }\\[10pt]
        {\makecell[l]{ {LfGP}\\ \citep{ablett2023learning}}}
        & IL
        & {$\checkmark$}
        & {$\checkmark$}
        & {$\times$}
        & {$\checkmark$}
        & {$\checkmark$}
        & {$\times$}
        & {$\times$ }\\[10pt]
        {\makecell[l]{MAPS\\ \citep{liu2023active}}}
        & IL$+$RL 
        & {$\checkmark$}
        & {$\checkmark$}
        & {$\checkmark$}
        & {$\checkmark$}
        & {$\checkmark$}
        & {$\checkmark$}
        & {$\times$ }\\[10pt]
        {\makecell[l]{\textbf{\algname~(Ours)}}}
        & IL$+$RL 
        & {$\checkmark$}
        & {$\checkmark$}
        & {$\checkmark$}
        & {$\checkmark$}
        & {$\checkmark$}
        & {$\checkmark$}
        & {$\checkmark$}\\
        \bottomrule
    \end{tabular}}}
\end{table*}

\subsection{\fix{Additional Notes on Related work}}
\fix{
MAMBA addressed the challenge of learning from multiple sub-optimal oracles and tackled two fundamental questions: what constitutes a reasonable benchmark for policy improvement, and how to systematically combine sub-optimal oracles into a stronger baseline. MAMBA proposed a max-aggregated baseline and suggested policy improvement from it as a natural strategy for combining these oracles to form a better policy. In addition, they introduced a novel Max-aggregation of Multiple Baseline approach and provided a theoretical performance guarantee for it. However, one limitation of MAMBA is its high sample complexity. It requires prolonged rounds to identify the optimal oracle for a given state due to its strategy of uniformly sampling an oracle, resulting in a larger accumulation of regret.
MAPS aims to enhance the sample efficiency of MAMBA by introducing Max-aggregation Active Policy Selection with theoretical support, and it empirically outperforms MAMBA. Nevertheless, both MAMBA and MAPS share a common limitation in non-robustness. They are susceptible to the quality of the oracle set, and both algorithms may fail in cases where the oracle set is of poor quality. 
Our work addresses this robustness challenge by proposing a novel $\maxplus$ framework. Inspired by the max-aggregation policy from MAMBA, we introduced a $\maxplus$-aggregation policy based on a novel extended oracle set. This enables the policy to undergo self-improvement even when the oracle set is poor. Additionally, we proposed a novel algorithm, RPI, with Robust Active Policy Selection to improve active policy selection from MAPS. Theoretical analyses were provided for both the $\maxplus$ framework and the RPI algorithm.}

\section{Additional Background}\label{app:additional_background}

\subsection{Additional algorithms for learning from multiple oracles}
In this section, we {introduce} a few baselines that learn from a set of black-box oracles $\Policies=\curlybracket{\policy^k}_{k\in\bracket{\ONum}}$.

\textbf{Single-best expert} $\policy^\star$: The first baseline that we consider imitates a single oracle that achieves the best performance in hindsight among the oracle set, i.e., $\pi^\star \coloneqq \argmax_{\pi \in \Pi} \mathbb{E}_{s_0 \sim d_0} \big[V^{\pi}(s_0)\big]$. After figuring out the single-best expert, this strategy simply keeps rolling out the expert. In practice, this is often inadequate as it neglects 
the potential benefits of suboptimal oracles at the state level.

\textbf{Max-following} $\policy^{\bullet}$: 
Given a collection of $k$ imitation learning \textit{oracles} $\OracleSet=\{\policy^k\}_{k\in\bracket{\ONum}}$, %
the \textit{max-following} policy \citep{cheng2020policy,liu2023active} is a greedy policy that selects the oracle with the highest expertise in any given state:
\begin{equation*}\label{eq:max_following}
    \policy^{\bullet}\paren{\action \mid \state}:=\policy^{k^\star}\paren{\action \mid \state}, \;\; k^\star:=\argmax_{k\in\bracket{\ONum}} \; \VFunc^{k}\paren{\state}
\end{equation*}
where $V^k(s) = V^{\pi^k}(s)$ is the value function for oracle $k \in [\ONum]$. %

\textbf{Max-aggregation} $\policy^\text{max}$: %
Max-aggregation~\citep{cheng2020policy} performs one-step improvement based on the \emph{max-following} policy $\policy^{\bullet}$. Denote a natural value baseline $\Func^\text{max}\paren{\state_t}$ for IL with multiple oracles as
\begin{equation}\label{eq:fmax}
    \Func^\text{max}\paren{\state_t}:=\max_{k\in \bracket{\ONum}}\VFunc^{k}\paren{\state}.
\end{equation}
We then denote the \emph{max-aggregation} policy as 
\begin{equation}\label{eq:maxagg}
\centering
\begin{split}
     \policy^\text{max}\paren{\action \mid \state} := \delta_{\action=\action^{\star}}, \textrm{where }& \action^{\star} = \argmax_{\action\in\actionSpace}\; \AFunc^{\Func^\text{max}}\paren{\state,\action},\\
    \AFunc^{\Func^\text{max}}\paren{\state,\action} = \RFunc\paren{\state,\action} + \expctover{\state'\sim\transDynamics | \state,\action}{\Func^{\textrm{max}}\paren{\state'}} &-\Func^{\textrm{max}}\paren{\state}, \textrm{and $\delta$ is the Dirac delta distribution.}        
\end{split}
\end{equation}
The max-aggregation policy is a function of $\Func^\text{max}$ and, in turn, requires knowledge of the MDP and each oracle's value function (Eqn.~\ref{eq:fmax}). However, in the episodic interactive IL setting, oracles are provided as black boxes and their value functions are unknown. MAMBA~\citep{cheng2020policy} and MAPS~\citep{liu2023active} deal with this by reducing IL to an online learning problem and adapt the online loss defined at {round} $n$ as:%
\begin{equation}\label{eq:mamba_lops_loss}
    \begin{split}
        \loss_n\paren{\policy;\lambda} := &-(1-\lambda)\horizon \expctover{\state\sim\stateDist^{{\policy_n}}}{\AFunc^{f^\textrm{max},\policy}_{\lambda}\paren{\state,\policy}} - \lambda \expctover{\state\sim\stateDist_0}{\AFunc_{\lambda}^{f^\textrm{max},\policy}\paren{\state,\policy}}.
    \end{split}
\end{equation}
Here, $\Adv^{f^\textrm{max},\policy}_{\lambda}(\state,\action)$ is a $\lambda$-weighted advantage defined as:
\begin{equation}\label{eq:lambda_weight_advantage}
    \Adv^{f^\textrm{max},\policy}_{\lambda}\paren{\state,\action}:= \paren{1-\lambda}\sum_{i=0}^{\infty}\lambda^{i}\Adv_{\paren{i}}^{f^\textrm{max}, \policy}\paren{\state,\action},
\end{equation}
which integrates various $i$-step advantages:
\begin{equation}\label{eq:i_step_advantage}
    \begin{split}
    \Adv^{f^\textrm{max},\policy}_{\paren{i}}\paren{\state_t,\action_t} := \;&\mathbb{E}_{\trajectory_t\sim \rho^\pi\paren{\cdot|\state_t}}
        {[\reward(\state_t,\action_t)} + \cdots +\reward\paren{\state_{t+i},\action_{t+i}}+\Func^\text{max}\paren{\state_{t+i+1}})]-\Func^\text{max}\paren{\state_t}. \notag
    \end{split}
\end{equation}

\subsection{Limitations of the prior art}
MAPS~\citep{liu2023active} and MAMBA~\citep{cheng2020policy} suffer from two limitations related to their non-robustness to the choice of the oracle set. First, their online loss function \eqref{eq:mamba_lops_loss} relies on a predetermined $\lambda$ value that combines imitation learning and reinforcement learning, making their performance sensitive to the {quality} of the oracle set. Second, their gradient estimator utilizes the max-aggregation policy $\policy^{\textrm{max}}$ and the value baseline function $\Func^\textrm{max}$, both of which are dependent on the given black-box oracle set. If the oracle set includes only adversarial oracles, these methods will still try to perform imitation learning, thereby impeding policy enhancement.

\subsection{Value function approximator for discrete environment}
In the interactive episodic MDP, we roll out a selected oracle $k$, resulting in $\ENum_{k}(\state_t)$ trajectories $\trajectory_{1,k},\trajectory_{2,k}, \ldots, \trajectory_{\ENum_k,k}$ starting from state $\state_t$ for round $N$. 
We determine an estimate for the return in state $\state_t$ by averaging the returns obtained across the trajectories:
\begin{equation}\label{eq:mu_v}
    \hat{\VFunc}^{\policy_k}\paren{\state_t}=\frac{1}{\ENum_k\paren{\state_t}}\sum_{i=1}^{\ENum_k\paren{\state_t}}\sum_{j}^{\horizon}\lambda^j\RewardFunc\paren{\state_j,\action_j}.
\end{equation}

\subsection{Active policy selection} 

To address the sample efficiency challenge in learning from multiple experts, we reference active policy selection technique in MAPS work to select the best oracle $k^{\star}$ for state $\state_t$ as follows:

\begin{equation}
    {k_\star}=\argmax_{k \in \bracket{\ONum}}
    \begin{cases}
         \hat{V}^k(s_{t}) + \sqrt{\frac{2\horizon^2\log{\frac{2}{\delta}}}{\ENum_k\paren{\state_{t}}}} &\stateSpace \, \textrm{discrete} \\
         \hat{V}_{\mu}^k(s_{t}) + \sigma_k\paren{\state_t} &\stateSpace \, \textrm{continuous}
    \end{cases}
\end{equation}

\section{Proofs}

In the following, we provide proofs for the theoretical claims in the main paper.

\begin{lemma}\citep{kakade2002approximately,ng1999policy}\label{lem:PDL}
    Let $\Func:\stateSpace \rightarrow \mathbb{R}$ such that $\Func\paren{\state_{ {\horizon}}}=0$. For any \textit{MDP} and policy $\policy$,
    \begin{equation}\label{lem:performance_difference_lemma}
        \VFunc^{\policy}\paren{\stateDist_0}-\Func\paren{\stateDist_0}= {\horizon}\expctover{\state\sim\stateDist^{\policy}}{\AFunc^{\Func}\paren{\state,\policy}}
    \end{equation}
\end{lemma}

From \lemref{lem:PDL}, we get the following corollary:
\begin{corollary}\citep{cheng2020policy}\label{col:performance_improvable}
    If $\Func$ is improvable with respect to $\policy$, then $\VFunc^{\policy}\paren{\state}\geq \Func\paren{\state}$, $\forall \state \in \stateSpace$.
\end{corollary}
\corref{col:performance_improvable} indicates that a policy $\policy$ outperforms all policies in $\ExtendOracleSet$, if, for every state, there is a baseline value function $\Func$ superior to that of all policies $(\Func (\state) \geq \VFunc^k\paren{\state}, \forall k \in [\lvert \ExtendOracleSet \rvert], \state \in \stateSpace)$, while $\Func$ can be improved by $\policy$~(i.e., $\AFunc^{\Func}\paren{\state,\policy}\geq 0$).

\subsection{Proof of \propref{pro:policy_oplus}}\label{pr:advantage}

\begin{proof}
    Without loss of generality, let us assume the optimal oracle is oracle 1 (the first oracle) in oracle set $\Policies$, 
    \begin{subequations}
        \begin{align}
            \aplus\paren{\state,\polmaxplusflw} &= \reward\paren{\state,\polmaxplusflw} + \expctoveronly{\action \sim \polmaxplusflw|\state}\expctoveronly{\state'\sim \transDynamics|\state,\action}\bracket{\fplus\paren{\state'}} - \fplus\paren{\state}\\
            &\geq \reward\paren{\state,\polmaxplusflw} + \expctoveronly{\action\sim \polmaxplusflw|\state}\expctoveronly{\state'\sim \transDynamics|\state,\action}\bracket{\VFunc^{1}\paren{\state'}} - \VFunc^{1}\paren{\state}\\
            &\geq \reward\paren{\state,\policy^{\bullet}} + \expctoveronly{\action\sim \polmaxplusflw|\state}\expctoveronly{\state'\sim \transDynamics|\state,\action}\bracket{\VFunc^{1}\paren{\state'}} - \VFunc^{1}\paren{\state}\\
            &=\AFunc^{\VFunc^{1}}\paren{\state,\policy^{1}}\geq 0,
        \end{align}
    \end{subequations}
    where the last step follows since $\polmaxplusflw\paren{\action|\state} \geq \policy^{\bullet}\paren{\action|\state}= \policy^{1}\paren{\action|\state}$. Since we have $\aplus\paren{\state,\polmaxplusflw}\geq 0$, together with \lemref{lem:PDL}, we have 
\begin{align}
 \VFunc^{\polmaxplusflw}(\state) \geq \fplus\paren{\state} = \max_{k\in\bracket{|\ExtendOracleSet|}}\VFunc^{k}(\state).
\end{align}    
$\VFunc^{\polmaxplusflw}(\state) \geq \fplus\paren{\state}$ indicates that 
following $\polmaxplusflw$ is equally good or superior to imitating a single best policy in $\ExtendOracleSet$.
\end{proof}

\subsection{Proof of \propref{prop:max_plus_performance_lowerbound}}

We denote $\fplus_0(\state) := \argmax_{k\in\bracket{\ONum}} \; \VFunc^{k}\paren{\state}$. 
According to Theorem 1 of \citet{cheng2020policy}, we obtain
\begin{align}
 \expct{\max_{n\in\bracket{\ENum}} \VFunc^{\policy_n} \paren{\stateDist_0}} \geq  \expctover{\state\sim\stateDist_0}{
        \fplus_0\paren{\state}
        } + \expct{\Delta_N - \epsilon_{\ENum}\paren{\Policyclass} - {\mathrm{Regret}^{\mathcal{L}}_N}} .  
\end{align}
{Now let $\Policies^{\mathcal{E}}_m=\OracleSet\cup \policy_m$. 
Following the same reasoning strategy as above, we will have lower bound for \algname as $\expctover{\state\sim\stateDist_0}{
        \fplus_m\paren{\state}
        } + \expct{\Delta_N - \epsilon_{\ENum}\paren{\Policyclass} - {\mathrm{Regret}^{\mathcal{L}}_N}}$.} Since $\expctover{\state\sim\stateDist_0}{\fplusAtM\paren{\state}}\geq \expctover{\state\sim\stateDist_0}{\fplus_0\paren{\state}}$, we have performance lower bound of \algname no worse than MAMBA. 

\textbf{Remark.} MAPS \citep{liu2023active} retains MAMBA's lower bound but enhances sample efficiency and reduces the bias in $\textrm{Regret}_N^{\mathcal{L}}$. The inherent uncertainty of the optimal policy $\policy^{\star} \in \Policies^{\mathcal{E}}_m$ makes an unbiased $\fplus$ estimate challenging. The regret term $\mathrm{Regret}^{\mathcal{L}}_N$ is bounded by: 
$$\mathbb{E}\bracket{\textrm{Regret}^{\mathcal{L}}_{\ENum}} \leq O\paren{\paren{\beta^{+} + \beta^{\epsilon}
}\ENum + \sqrt{\var\ENum}},$$ where $\beta^{+}$ is the estimation bias that results from selecting the non-optimal policy $\hat{\policy}^{\star}$ in $\Policies^{\mathcal{E}}_m$
for a given state, and 
$\beta^{\epsilon}$ is the value estimation error w.r.t.\ the true value for given state of selected policy $\hat{\policy}^{\star}$ and $\var$ represents the variance term. 

MAPS improves upon MAMBA's sample complexity, reducing bias in its regret bound via an active policy selection mechanism. Our work builds on MAPS, emphasizing empirical enhancements in active policy selection with the integration of the learner policy in $\ExtendOracleSet_m$. %

\section{Experimental Details}\label{app:experiment}

\subsection{Baselines}\label{app:experiment:baselines}

\paragraph{AggreVaTeD}
AggreVaTeD~\citep{sun2017deeply} is a differentiable version of AggreVaTe, which focuses on a single oracle scenario. AggreVaTeD allows us to train policies with efficient gradient update procedures. AggreVaTeD models the policy as a deep neural network and trains the policy using differentiable imitation learning. By applying differentiable imitation learning, it minimize the difference between the expert's demonstration and the learner policy behavior. AggreVaTeD learns from the expert's demonstration while interact with the environment to outperform the expert.

\paragraph{Max-Aggregation} 
We have developed a variant of the Max-aggregation policy as outlined in Equation~\eqref{eq:maxagg} that is specifically designed for pure imitation learning using multiple oracle sets. When utilizing a single oracle, it effectively reduces to AggreVateD. Our approach builds on the existing MAMBA framework by setting the lambda value in the loss function to zero. While max-aggregation may not always yield the optimal policy, it offers the advantage of being able to achieve results with fewer samples, making it a more sample-efficient option.

\paragraph{ {LOKI-variant}}
LOKI~\citep{cheng2018fast} is strategy for policy learning that combines the imitation learning and reinforcement learning objectives in a two-stage manner for the single oracle setting. In the first stage, LOKI performs imitation learning for a small but random number of iterations and then switches to policy gradient reinforcement learning method for the second stage.  LOKI is able to outperform a sub-optimal expert and converge faster than running policy gradient from scratch. In this work, we propose a variation of LOKI that adapts to multiple experts. During the first-half of training (i.e., the first stage) we perform Max-aggregation style imitation learning, and then perform pure reinforcement learning as the second stage.

\paragraph{PPO-GAE} \citet{schulman2015high} proposed the generalized advantage estimator (GAE) as a means of solving high-dimensional continuous control problems using reinforcement learning. GAE is used to estimate the advantage function for updating the policy. 
The advantage function measures how much better a particular action is compared to the average action. Estimating the advantage function with accuracy in high-dimensional continuous control problems is challenging. In this work, we propose PPO-GAE, which combines PPO's policy gradient method with GAE's advantage function estimate, which is based on a linear combination of value function estimates. By combining the advantage of PPO and GAE, PPO-GAE~\citep{schulman2017proximal} achieved both sample efficiency and stability in high-dimensional continuous control problems.

\paragraph{MAMBA}
MAMBA~\citep{cheng2020policy} is the SOTA work of learning from multiple oracles. It utilizes a mixture of imitation learning and reinforcement learning to learn a policy that is able to imitate the behavior of multiple experts.  MAMBA is also considered as interactive imitation learning algorithm, it imitates the expert and interact with environment to improve the performance. MAMBA randomly select the state as switch point between learner policy and oracle. Then, it randomly selects the oracle to roll out. It effectively combines the strengths of multiple experts and able to handle the case of conflicting expert demonstrations.

\paragraph{MAPS} 
MAPS~\citep{liu2023active} is a policy improvement algorithm that performs imitation learning from multiple suboptimal oracles. It actively chooses the oracle to imitate based on their value function estimates and identifies the states that require exploration. By introducing two variations, Active Policy Selection (APS) and Active State Exploration (ASE), MAPS improves the sample efficiency of MAMBA. The MAPS variant selects the most promising oracle, denoted as $k_{\star}$, for rollout, utilizing the resulting trajectory to refine the value function estimate $\hat{\VFunc}^{k_{\star}}\paren{\state_t}$. This approach aims to minimize the chances of selecting an inferior oracle for a given state $\state_t$, thereby reducing both the sample complexity and gradient estimation bias. On the other hand, the ASE variant of MAPS deliberates whether to continue with the current policy or switch to what is believed to be the most promising oracle, similar to APS, by leveraging an uncertainty measure over the current state. In this study, we adopt MAPS variant as our baseline method.

\subsection{{Gym environments}}

We evaluate \algname and compare its performance to the aforementioned baselines on the Cheetah-run, CartPole-swingup, Pendulum-swingup, and Walker-walk tasks from the DeepMind Control Suite~\citep{tassa2018deepmind}  {and Window-close, Faucet-open, Drawer-close and Button-press from Meta-World \citep{yu2020meta}. In addition, we conduct experiments on a modified sparse reward Meta-World environment, which is considered to be a more challenge task.}

\subsection{Setup}
{\textbf{Setup.} In order to ensure a fair evaluation, all baselines are assessed using an equal number of environment interaction steps \fix{(training steps)}. Each training iteration involved a policy rollout for the same number of steps. We note that there is a discrepancy in the amount of data available to the learners of \algname and PPO-GAE. \algname (MAPS, MAMBA, Max-Aggregation) uses some of the interactions to learn the value function for each Oracle, which results in relatively less data for its learner, whereas PPO-GAE utilizes all the environment interactions to update all benefits for its learner policy. Thus, in this work, we balance the transition buffer size for each algorithm to make them have approximately same number of stored transitions for learner policy improvement. We average the result based on 5-10 trials.}

\subsection{Implementation details of \algname}

\begin{algorithm}[!ht]
    \caption{Robust Policy Improvement (\algname)} \label{alg:rpi:implement}
    \begin{algorithmic}[1] 
        \Require Learner policy $\policy_{1}$, oracle set $\Policies=\curlybracket{\policy^{k}}_{k\in \bracket{\ONum}}$, function approximators $\{\hat{\VFunc}^k\}_{k\in\bracket{\ONum}}$, $\hat{\VFunc}_n$
        \Ensure {The best policy in \curlybracket{\policy_1,...,\policy_{\ENum}}.}
        \For{$n=1,2, \ldots, N-1$} 
        \State Construct an extended oracle set $\ExtendOracleSet = \bracket{\policy^1, \policy^2, \ldots ,\policy^k, \policy_n}_{k\in \bracket{|\Policies|}}$.
        \State Sample $t_e\in \bracket{\horizon-1}$ uniformly random. We have a buffer with a fixed size ($|\mathcal{D}_n| = 19,200$) for each oracle, and we discard the oldest data when it fills up.
        \State Switch to $\policy^{k_{\star}}$
        \State Roll-in $\policy_n$ up to $t_e$, select $k_{\star}$ \eqref{eq:kstar}, and roll out $\policy^{k_{\star}}$ to collect data $\mathcal{D}^k$. \label{lin:aps:rollin:oracle}        
        \State Update $\hat{V}^{k_{\star}}$ using $\mathcal{D}^k$.
        \State Roll-out the learner policy $\policy_n$ for a specified steps ($2,048$), and add them to the buffer $\mathcal{D}_n'$.\label{lin:aps:rollin:learner} %
        \State Update $\hat{V}_n$ using $\mathcal{D}_n'$.
        
        \State Compute advantage  {$\estaplusGAE$}~\eqref{eq:A_Plus_GAE} and gradient estimate $\hat{\gradient}_n$~\eqref{eq:rpi_gradient} using $\mathcal{D}_n'$.

        \State Perform PPO style policy update on policy $\policy_n$ to $\policy_{n+1}$.\label{lin:aps:update}
        \EndFor 
    \end{algorithmic}
\end{algorithm}

We provide the details of $\algname$ in \algoref{alg:rpi} as \algoref{alg:rpi:implement}. \algoref{alg:rpi:implement} closely follows \algoref{alg:rpi} with a few modifications as follows:
\begin{itemize}
    \item In line~\ref{lin:aps:rollin:oracle}, we use a buffer with a fixed size ($|\mathcal{D}_n| = 19,200$) for each oracle, and discard the oldest data when it fills up.
    \item In line~\ref{lin:aps:rollin:learner}, we roll-out the learner policy until the buffer reaches a fixed size ($|\mathcal{D}_n'|$ = 2,048), and then empty it once we use the roll outs to update the learner policy. This stabilizes the training compared to storing a fixed number of trajectories in the buffer, as MAMBA does.
    \item In line~\ref{lin:aps:update}, we use PPO with a \maxplus~actor-critic style policy update.
    \item We pretrain the value function $\hat{V}^k$ of oracle $k$ before the main training loop, with trajectories generated by rolling out $\pi^k$ from the initial states. In the main training loop, we train $\hat{V}^k$ using the corresponding rolled-out trajectories, bootstrapped only by itself. This is the same strategy as in MAMBA and MAPS. Similarly, we train the learner value function $\hat{V}_n$  using only the trajectories rolled-in with $\pi_n$, bootstrapped only by itself.
\end{itemize}

\subsection{\fix{Hyperparameters and architectures}}
Table \ref{app:tab:hyperparams} provides a list of hyperparameter settings we used for our experiments.
We use an ensemble of MLPs to predict an oracle's value. 
With five identical MLPs that are separately initialized at random, we train each MLP separately for the same dataset.
At inference time, we collect predictions from all MLPs for a single input and compute the mean and standard deviation.

\begin{table*}[h!]
    \fix{
    \centering
    {\scriptsize
    \scalebox{1}{
    \begin{tabular}{c c }
        \toprule
        \textbf{Parameter} &  \textbf{Value} 
        \\
        \midrule
        {\makecell[l]{Shared}}
        \\
        \midrule
        {\makecell[l]{\quad Learning rate}} &  \makecell[c]{$3 \times 10^{-4}$}
        \\
        \midrule
        {\makecell[l]{\quad Optimizer}} &  \makecell[c]{Adam}
        \\
        \midrule
        {\makecell[l]{\quad Nonlinearity}} &  \makecell[c]{ReLU}
        \\
        \midrule
        {\makecell[l]{\quad \# of functions in a value function ensemble}} &  \makecell[c]{$5$}
        \\
        \midrule
        {\makecell[l]{Oracle}}
        \\
        \midrule
        {\makecell[l]{\quad \# of oracles in the oracle set $\paren{\ONum}$}} &  \makecell[c]{$3$}
        \\
        \midrule %
        {\makecell[l]{\quad The buffer size for oracle $k$ $\paren{|\mathcal{D}^k|}$}}  &  \makecell[c]{$19200$}
        \\
        \midrule %
        {\makecell[l]{\quad \# of episodes to rollout the oracles for value function pretraining}}  &  \makecell[c]{8}
        \\        
        \midrule
        {\makecell[l]{Learner}}
        \\
        \midrule
        {\makecell[l]{\quad Horizon of MetaWorld and DMControl $\paren{\horizon}$}}  &  \makecell[c]{$300$,~$1000$}
        \\
        \midrule %
        {\makecell[l]{\quad Replay buffer size for the learner policy $\paren{|\mathcal{D}_n'|}$}}  &  \makecell[c]{$2048$}
        \\
        \midrule %
        {\makecell[l]{\quad GAE gamma $\paren{\gamma}$}}  &  \makecell[c]{0.995}
        \\        
        \midrule %
        {\makecell[l]{\quad GAE lambda $\paren{\lambda}$ for AggreVaTeD and Max-Aggregation }}  &  \makecell[c]{0}
        \\
        {\makecell[l]{\quad \hphantom{GAE lambda $\paren{\lambda}$} for LOKi-variant}}  &  \makecell[c]{$0$ or $1$}
        \\
        {\makecell[l]{\quad \hphantom{GAE lambda $\paren{\lambda}$} for RPI}}  &  \makecell[c]{0.9}
        \\
        \midrule %
        {\makecell[l]{\quad \# of training steps (rounds) $\paren{\ENum}$}}  &  \makecell[c]{$100$}
        \\
        \midrule %
        {\makecell[l]{\quad \# of episodes to perform RIRO (Alg \ref{alg:rpi}, line 4) per training iteration}}  &  \makecell[c]{4}
        \\
        \midrule %
        {\makecell[l]{\quad mini-batch size}}  &  \makecell[c]{128}
        \\  
        \midrule
        {\makecell[l]{\quad \# of epochs to perform gradient updates per training iteration}}  &  \makecell[c]{4}
        \\        

        \bottomrule
    \end{tabular}}}
        \caption{{{RPI Hyperparameters}}. }
        \label{app:tab:hyperparams}
        }
\end{table*}

\subsection{Computing infrastructure \fix{and wall-time comparison}}
We conducted our experiments on a cluster that includes CPU nodes (approximately 280 cores) and GPU nodes (approximately 110 Nvidia GPUs, ranging from Titan X to A6000, set up mostly in 4- and 8-GPU configurations). \fix{Based on the computing infrastructure, we obtained the wall-time comparison in \tabref{table:wall-time} as follows.}

 \begin{table*}[ht!]
 \fix{
    \centering
    {\scriptsize
    \scalebox{1}{
    \begin{tabular}{l c c c c c c }
        \toprule
        \textbf{Methods}
        & {\makecell[c]{Cheetah}}
        & \makecell[c]{Cartpole}
        & \makecell[c]{Walk-Walker}
        & \makecell[c]{Pendulum}
        \\
        \midrule
        \textbf{\makecell[l]{MAPS}}
        &  \makecell[r]{1h 18m }
        &  \makecell[r]{1h 10m }
        &  \makecell[r]{1h 41m } 
        &  \makecell[r]{1h 17m } 
        \\
        \textbf{\makecell[l]{MAMBA}}
        &  \makecell[r]{1h 23m }
        &  \makecell[r]{59m }
        &  \makecell[r]{2h 14m }
        &  \makecell[r]{1h 21m }
        \\
        \textbf{\makecell[l]{LOKI-variant}}
        &  \makecell[r]{1h 17m }
        &  \makecell[r]{1h 36m }
        &  \makecell[r]{2h 11m }
        &  \makecell[r]{1h 12m }
        \\
        \textbf{\makecell[l]{PPO-GAE}}
        &  \makecell[r]{54m }
        &  \makecell[r]{58m }
        &  \makecell[r]{1h 10m }
        &  \makecell[r]{49m }
        \\
        \textbf{\makecell[l]{MAX-aggregation}}
        &  \makecell[r]{1h 5m }
        &  \makecell[r]{1h 34m }
        &  \makecell[r]{2h 25m }
        &  \makecell[r]{1h 13m }
        \\
         \midrule
        \textbf{\makecell[l]{RPI}}
        &  \makecell[r]{57m }
        &  \makecell[r]{58m }
        &  \makecell[r]{1h 43m }
        &  \makecell[r]{1h 18m }
        \\
        \bottomrule
    \end{tabular}}}
    \caption{{Wall-time comparison between different methods.} }
        \label{table:wall-time}   
        }
\end{table*}

\section{{Supplemental Experiments}}
\subsection{Comparing RPI against baselines with a data advantage}\label{app:baseline_alignment}
{In the main paper, we followed an experimental setup where we assumed that the learner had access to approximately the same quantity of transitions, while lacking access to the oracle's offline trajectory. However, some of the baseline algorithms, such as MAPS \citep{liu2023active} were originally evaluated under a different setting in the literature: They assume that the learner can access additional data from the oracles' pre-trained offline dataset. In this section,
we run MAPS under such a setting, in order to provide more comprehensive evaluation that is consistent with the literature. 
Note that under this experimental setup, MAPS has approximately twice the amount of data compared to RPI. We refer to this variant as MAPS-ALL. %

In contrast to MAPS in our original configuration, the performance of MAPS-ALL doubles in the Cheetah environment (as shown in Figure \ref{app:fig:results:cheetah:2}) and the Pendulum environment (as depicted in Figure \ref{app:fig:results:pendulum:2}). Moreover, the performance of MAPS-ALL surges between middle and end of rounds in the Pendulumn environment. This behavior mirrors what was reported in the original MAPS paper as well. 
In the Walker-walk environment (as illustrated in Figure \ref{app:fig:results:walker:2}), MAPS-ALL demonstrates an approximate 10\% improvement. For the Cartpole environment (Figure \ref{app:fig:results:cartpole:2}), MAPS-ALL's performance increases by around 20\%. MAPS-ALL exhibits overall performance similar to that of its original paper, with any differences caused by the difference in the oracle set. Notably, as a result, MAPS-ALL distinctly outperforms RPI only in the Pendulum environment. RPI's performance remains comparable to MAPS-ALL in the Cheetah environment and significantly surpasses the MAPS-ALL baseline in the Walker-Walker and Cartpole environments, despite utilizing much less data.
}
\begin{figure*}[ht!]
    \begin{subfigure}{.24\textwidth}
        \centering
        \includegraphics[%
        width=3.4cm,  clip={0,0,0,0}]{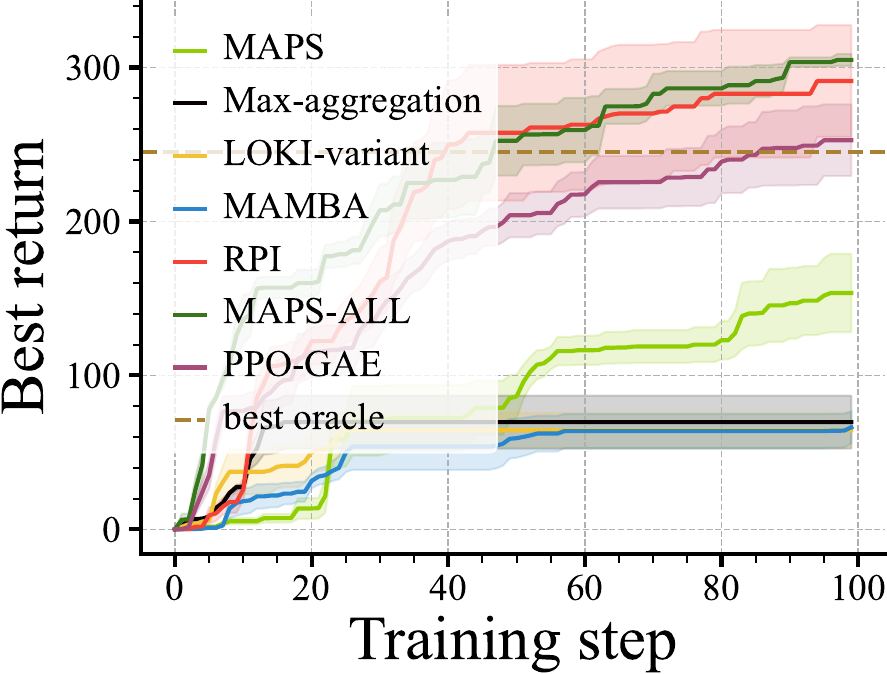}
        \caption{Cheetah (dense)}\label{app:fig:results:cheetah:2}
    \end{subfigure}\hfil
    \begin{subfigure}{.24\textwidth}
        \centering
        \includegraphics[%
        width=3.4cm,  clip={0,0,0,0}]{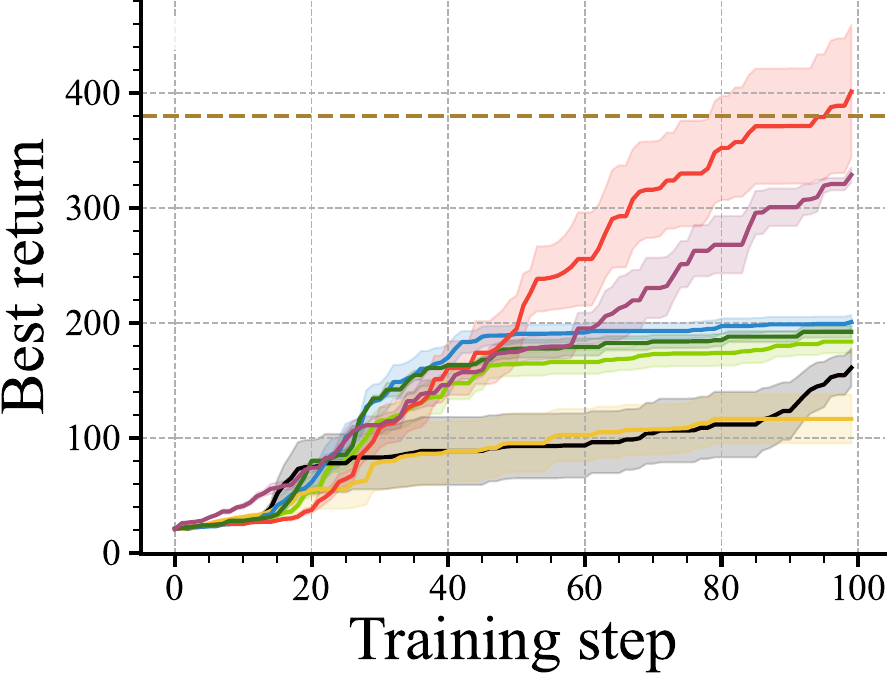}
        \caption{Walker-walk (dense)}\label{app:fig:results:walker:2}
    \end{subfigure}\hfil
    \begin{subfigure}{.24\textwidth}
        \centering
        \includegraphics[%
        width=3.4cm,  clip={0,0,0,0}]{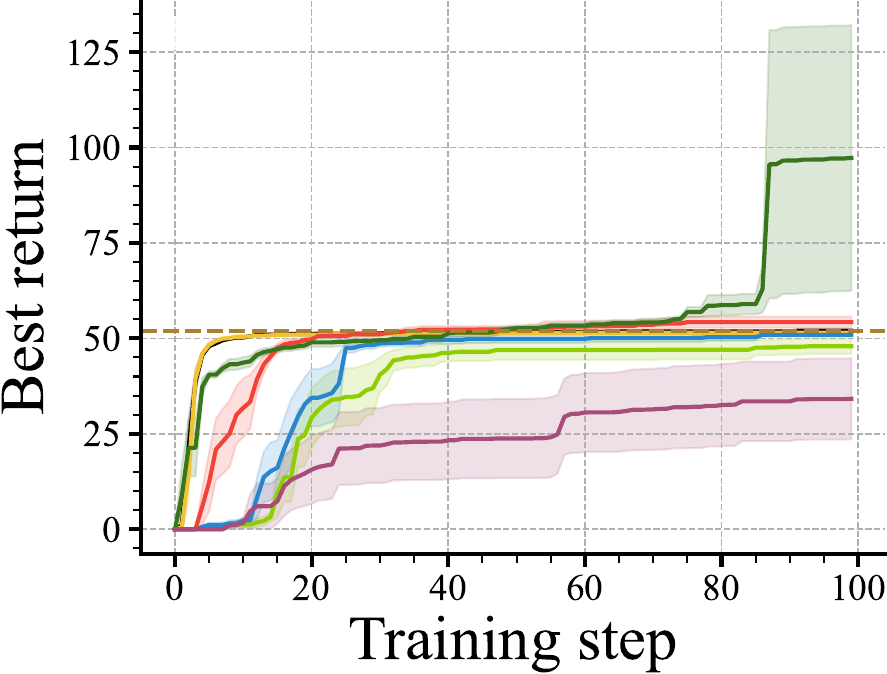}
        \caption{Pendulum (sparse)}\label{app:fig:results:pendulum:2}
    \end{subfigure}\hfil
    \begin{subfigure}{.24\textwidth}
        \centering
        \includegraphics[%
        width=3.4cm,  clip={0,0,0,0}]{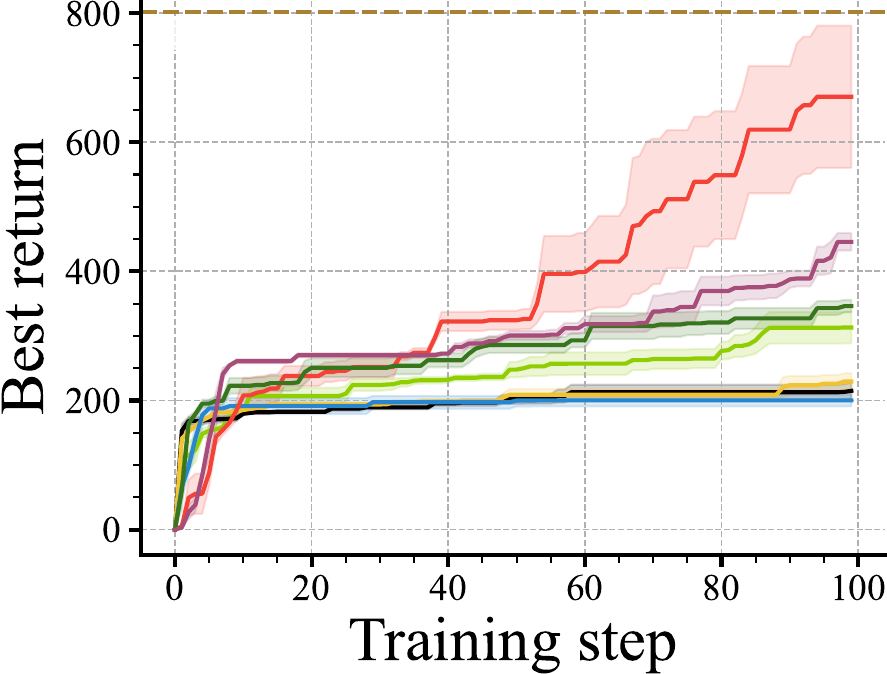}
        \caption{Cartpole (dense)}\label{app:fig:results:cartpole:2}
    \end{subfigure}\hfil

    \caption{Running MAPS in the original paper's setting.}
    
  \label{app:fig:exp:multi_experts_align}
\end{figure*}

\subsection{{ Meta-World experiments (dense reward)}}\label{sec:metaworld-dense}
In \figref{fig:exp:dense:meta:app}, we conducted  additional experiments comparing RPI and state-of-the-art (SOTA) methods \{MAMBA, MAPS\}, as well as the best-performing oracle and PPO-GAE, across the Meta-World benchmarks. The tasks are including (1) \texttt{window-close}, (2) \texttt{faucet-open}, (3) \texttt{drawer-close}, and (4) \texttt{button-press}. RPI demonstrates superior performance compared to all baselines in the majority of environments, with the exception of the button-press task.

\begin{figure*}[ht!]
    \begin{subfigure}{.24\textwidth}
        \centering
        \includegraphics[%
        width=3.4cm,  clip={0,0,0,0}]{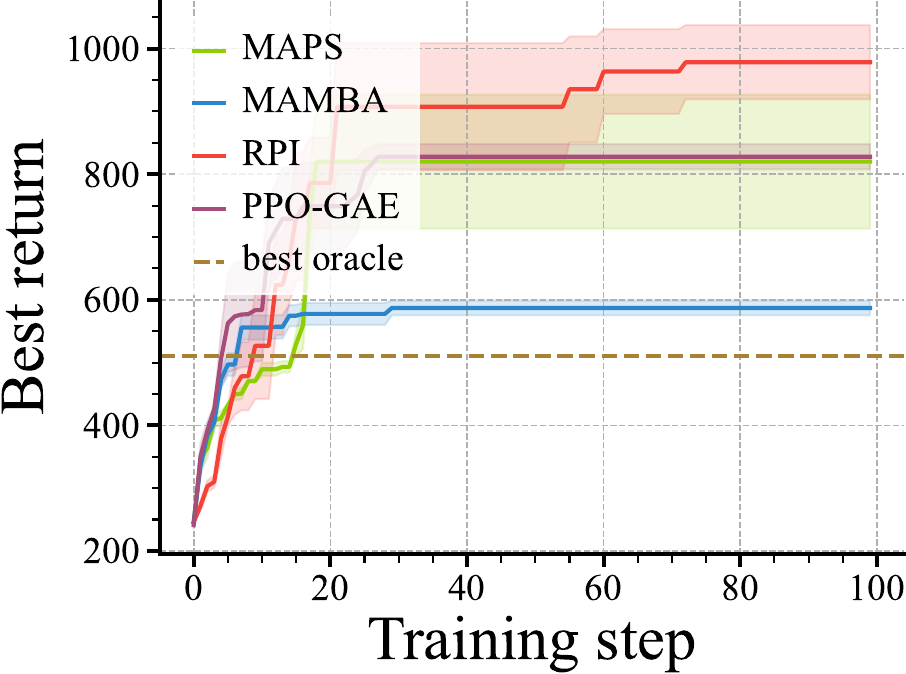}
        \caption{Window-close (dense)}\label{fig:results:window-close:3}
    \end{subfigure}\hfil
    \begin{subfigure}{.24\textwidth}
        \centering
        \includegraphics[%
        width=3.4cm,  clip={0,0,0,0}]{figures/metaworld-faucet-open-plot-p7p8p9-faucet-open-0.pdf}
        \caption{Faucet-open (dense)}\label{fig:results:faucet-open:3}
    \end{subfigure}\hfil
    \begin{subfigure}{.24\textwidth}
        \centering
        \includegraphics[%
        width=3.4cm,  clip={0,0,0,0}]{figures/metaworld-drawer-plot-p4p5p6-drawer-close-0.pdf}
        \caption{Drawer-close (dense)}\label{fig:results:drawer-close:3}
    \end{subfigure}\hfil
    \begin{subfigure}{.24\textwidth}
        \centering
        \includegraphics[%
        width=3.4cm,  clip={0,0,0,0}]{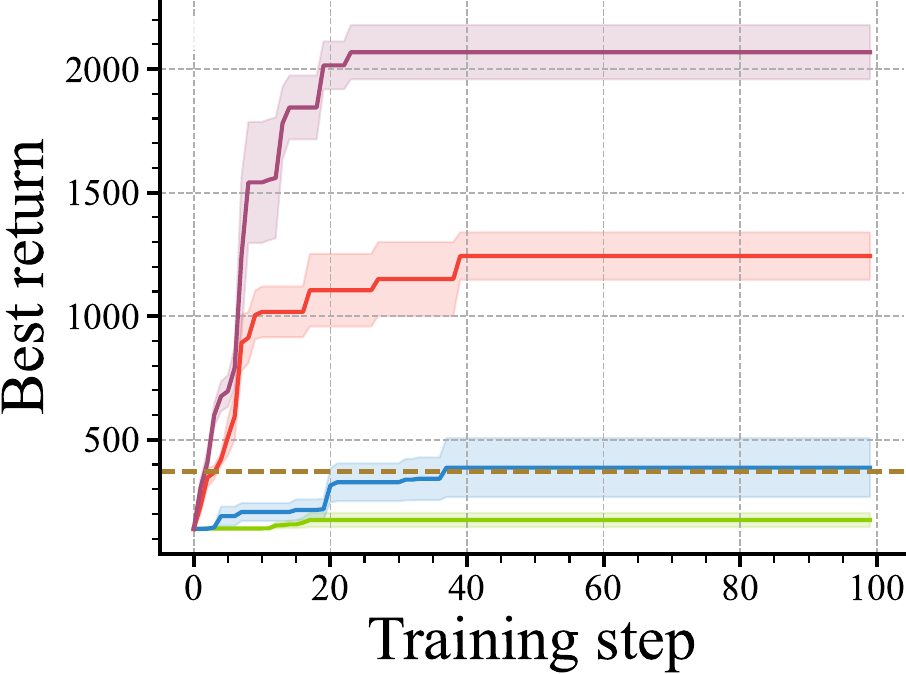}
        \caption{Button-press (dense)}\label{fig:results:button-press:3}
    \end{subfigure}\hfil
    \caption{ Experimental results on the Meta-World benchmark with dense reward.}
    
  \label{fig:exp:dense:meta:app}
\end{figure*}

\subsection{ {Meta-World experiments (sparse reward)}}\label{sec:metaworld-sparse}
In \figref{fig:exp:meta:app:sparse}, to further demonstrate the advantages of imitation learning, we modified the Meta-World environment to create a more challenging sparse reward environment. In this environment, the agent only receives a reward of 1 upon success; otherwise, it receives a reward of 0. We then compared the performance of the RPI and state-of-the-art (SOTA) imitation learning-based methods {MAMBA, MAPS}, as well as the pure RL method PPO-GAE, and the best-performing oracle across the Meta-World benchmarks. The tasks include (1) \texttt{window-close}, (2) \texttt{faucet-open}, (3) \texttt{drawer-close}, and (4) \texttt{button-press}. In these sparse reward environments, when provided with a good oracle, the imitation learning-based approach demonstrates its advantage over the pure RL approach. RPI, MAPS, and Mamba outperform PPO-GAE by a factor of 3 in the \texttt{button-press} environment. When provided with a bad oracle, RPI can still outperform MAPS and MAMBA in the \texttt{faucet-open} environment. Moreover, even with a poor oracle, RPI outperforms both IL-based approaches (MAMBA, MAPS) and the RL-based approach (PPO-GAE) in the \texttt{window-close} environment, showcasing that RPI enjoys benefits from both RL and IL aspects.

\begin{figure*}[ht!]
    \begin{subfigure}{.24\textwidth}
        \centering
        \includegraphics[%
        width=3.4cm,  clip={0,0,0,0}]{figures/metaworld-window-sparse-plot-p2p4p6-window-close-0.pdf}
        \caption{Window-close (sparse)}\label{fig:results:window-close:sparse}
    \end{subfigure}\hfil
    \begin{subfigure}{.24\textwidth}
        \centering
        \includegraphics[%
        width=3.4cm,  clip={0,0,0,0}]{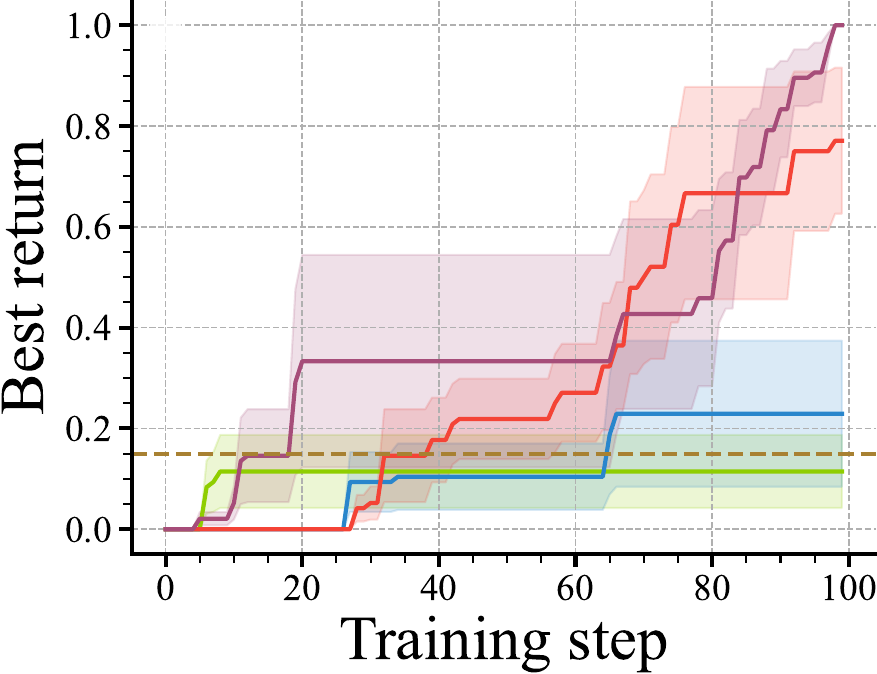}
        \caption{Faucet-open (sparse)}\label{fig:results:faucet-open:sparse}
    \end{subfigure}\hfil
    \begin{subfigure}{.24\textwidth}
        \centering
        \includegraphics[%
        width=3.4cm,  clip={0,0,0,0}]{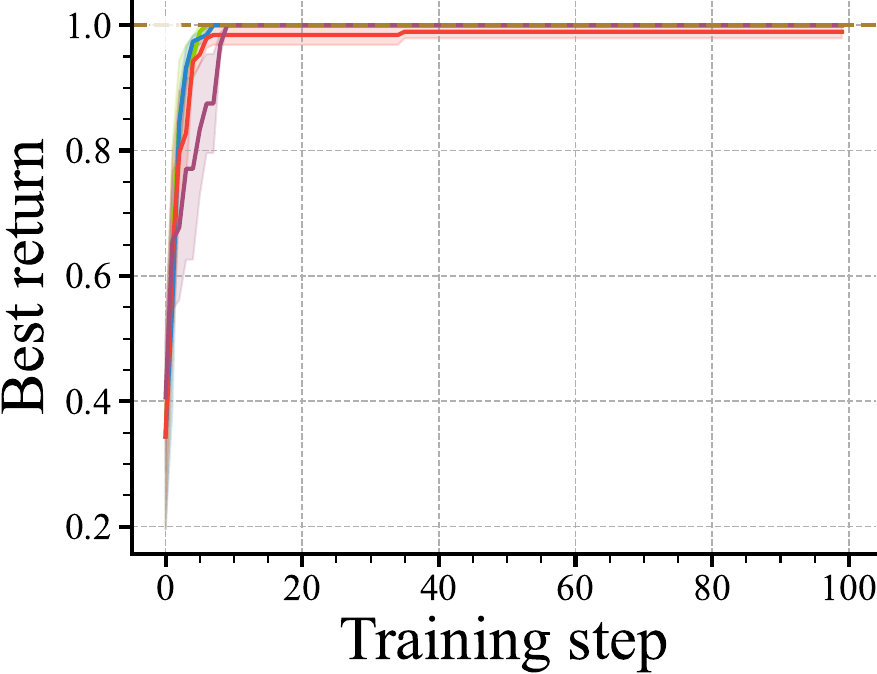}
        \caption{Drawer-close (sparse)}\label{fig:results:drawer-close:sparse}
    \end{subfigure}\hfil
    \begin{subfigure}{.24\textwidth}
        \centering
        \includegraphics[%
        width=3.4cm,  clip={0,0,0,0}]{figures/metaworld-button-sparse-plot-p2p2p3-button-press-0.pdf}
        \caption{Button-press (sparse)}\label{fig:results:button-press:sparse}
    \end{subfigure}\hfil
    \caption{ Experimental results on the Meta-World benchmark with sparse reward.}
    
  \label{fig:exp:meta:app:sparse}
\end{figure*}

\subsection{{Single/empty oracle set}}
In \figref{fig:exp:multi_experts}, we mainly discuss experiment on multiple oracle set. In \figref{fig:exp:ablation:single:empty:set}, we demonstrate that \algname is also robust enough to handle single or empty oracle set.

\textbf{Single oracle setting.} In \figref{fig:ablation:single_oracle}, we demonstrate that \algname outperforms all other baselines in a single oracle setting as well. %
This is consistent with the results observed in the multiple-experts setting. %
\figref{fig:results:helpful_oracle} demonstrates that providing an oracle with mediocre performance to \algname boosts the performance rather than providing a near-optimal oracle. Since we train oracles' value functions from the oracle rollouts, the value function of the near-optimal oracle may not have seen the ``bad'' states that the learner policy would encounter in the early stage. This leads to inaccuracy in the predicted values for such states. In comparison, the value function for a mediocre oracle would be able to produce accurate predictions on such states. %

\textbf{No oracle environment.} When there are no experts available, the performance of imitation learning-based approaches will inevitably degrade. However, as shown in \figref{fig:ablation:no_expert}, \algname can adapt to such a scenario by regressing to pure reinforcement learning. 
Since we extend \algname based on PPO-GAE, it achieves a similar level of performance to PPO-GAE when the oracle set is empty.

\begin{figure*}[ht!]
\begin{subfigure}{.32\textwidth}
    \centering
    \includegraphics[%
    width=3.4cm,  clip={0,0,0,0}]{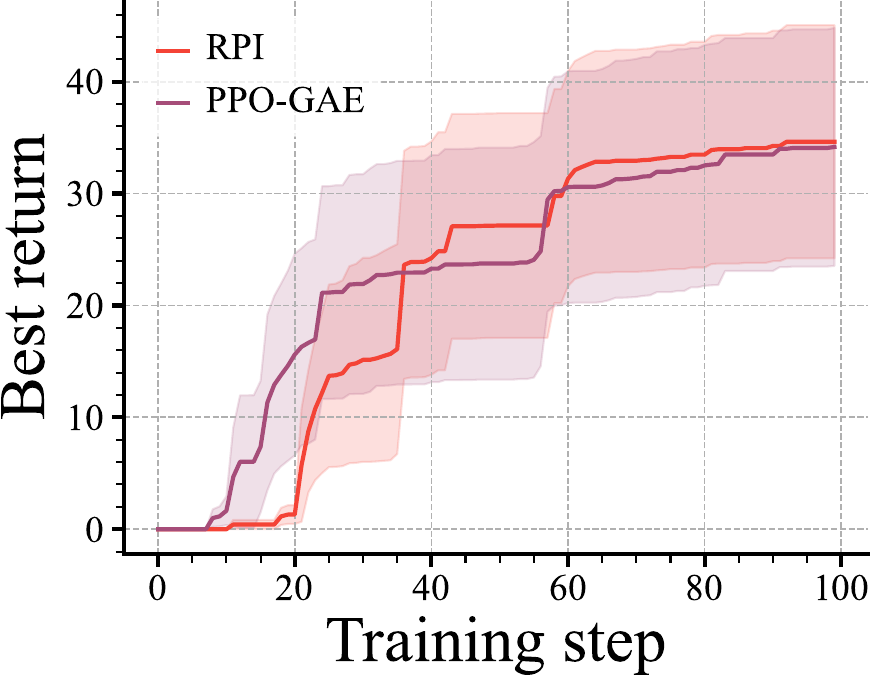}
    \caption{No oracles}\label{fig:ablation:no_expert}
\end{subfigure}
\begin{subfigure}{.32\textwidth}
    \centering
    \includegraphics[%
    width=3.4cm,  clip={0,0,0,0}]{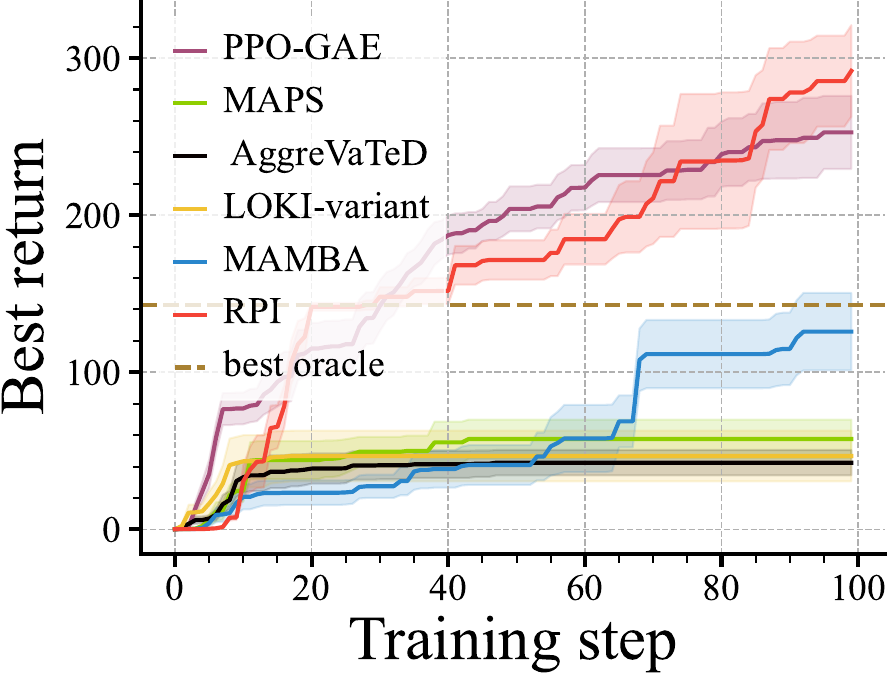}
    \caption{Good single oracle}\label{fig:ablation:single_oracle}
\end{subfigure}
\begin{subfigure}{.32\textwidth}
    \centering
    \includegraphics[%
    width=3.4cm,  clip={0,0,0,0}]{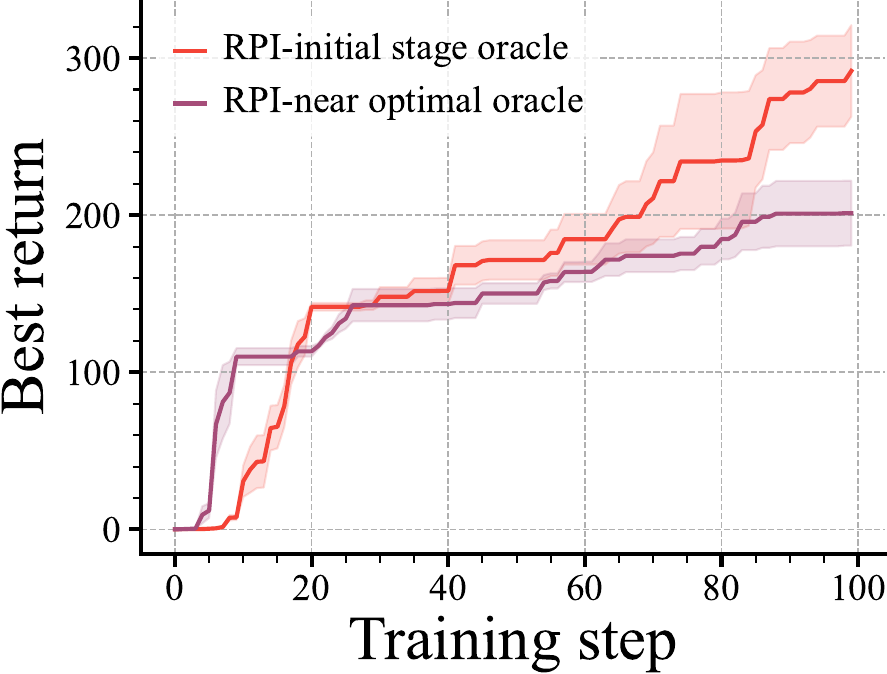}
    \caption{Helpful oracle ablation}\label{fig:results:helpful_oracle}
\end{subfigure}
    \centering
        \caption{\textbf{Ablation study.} 
        (a) Comparing \algname and PPO-GAE under no oracle Pendulum environment. 
        (b) Comparing \algname and six baseline under single expert Cheetah environment. 
        (c) ablation study on oracle quality under Cheetah environment. 
        }
  \label{fig:exp:ablation:single:empty:set}
\end{figure*}

\subsection{ {Ablation on confidence threshold $\threshold$}} \label{app:tuning_gamma}

In \tabref{app:tab:threshold:ablation}, we perform an ablation study of threshold $\threshold$ in \eqnref{eq:f_plus_confidence_aware} of robust policy gradient component.
We treat $\threshold$ as a hyperparameter the choice of which depends on the user's risk aversion.
Empirically, we find that a setting of $\threshold=0.5$ robustly works well in nearly every setting in our experiments across Deepmind Control suite, Meta-World (Dense reward), and Meta-world (Sparse reward) environment, the one exception being the Pendulum domain. To better understand how the choice of $\threshold$ effects overall performance, we conducted a set of experiments in which we ran RPI on the Deepmind Control Suite using different values for $\threshold\in \bracket{0,0.5,1,3,5}$. As the following table shows, setting $\threshold=0.5$ yields the best performance for all but the Pendulum environment.

In practice, one can first use roll outs of each oracle to estimate the standard deviation and associated confidence intervals of their ensemble values. A conservative user could then start by setting $\threshold$ based on a probabilistic lower-bound of $\sigma$ and subsequently tune the hyperparameter according to user's risk aversion preference.

\begin{table*}[h!]
    \centering
    {\scriptsize
    \scalebox{1}{
    \begin{tabular}{l c c c c c c }
        \toprule
        \textbf{Environment}
        & {\makecell[c]{Round}}
        & \makecell[c]{$\threshold=0$}
        & \makecell[c]{$\threshold=0.5$}
        & \makecell[c]{$\threshold=1$}
        & \makecell[c]{$\threshold=3$}
        & \makecell[c]{$\threshold=5$}
        \\
        \midrule
        \textbf{\makecell[l]{Cheetah}}
        &  \makecell[c]{100}
        &  \makecell[c]{252.7 ± 23.2 }
        &  \makecell[c]{\textbf{291.2} ± 36.3} 
        &  \makecell[c]{251.4 ± 15.1}
        &  \makecell[c]{53.4 ± 20.0}
        &  \makecell[c]{81.3 ± 20.8}
        \\
        \midrule
        \textbf{\makecell[l]{Walker-walk}}
        &  \makecell[c]{100}
        &  \makecell[c]{328.7 ± 6.5}
        &  \makecell[c]{\textbf{402.2} ± 57.7}
        &  \makecell[c]{253.0 ± 43.5}
        &  \makecell[c]{31.8 ± 1.4}
        &  \makecell[c]{38.2 ± 1.9}
        \\
        \midrule
        \textbf{\makecell[l]{Pendulum}}
        &  \makecell[c]{100}
        &  \makecell[c]{34.2 ± 23.5}
        &  \makecell[c]{38.0 ± 10.4}
        &  \makecell[c]{45.6 ± 2.3}
        &  \makecell[c]{\textbf{54.3} ± 1.5}
        &  \makecell[c]{52.1 ± 0.1}
        \\
        \midrule
        \textbf{\makecell[l]{Cartpole}}
        &  \makecell[c]{100}
        &  \makecell[c]{445.7 ± 13.5}
        &  \makecell[c]{\textbf{670.4} ± 110.1}
        &  \makecell[c]{394.8 ± 50.6}
        &  \makecell[c]{301.7 ± 60.0}
        &  \makecell[c]{303.2 ± 4.0}
        \\
        \bottomrule
    \end{tabular}}}
        \caption{{Tuning the confidence threshold $\threshold$}. }
        \label{app:tab:threshold:ablation}
\end{table*}

\subsection{{Ablation on UCB/LCB policy selection}}\label{app:lcb-ucb-raps}

\begin{figure*}[ht!]
    \begin{subfigure}{.24\textwidth}
        \centering
        \includegraphics[%
        width=3.4cm,  clip={0,0,0,0}]{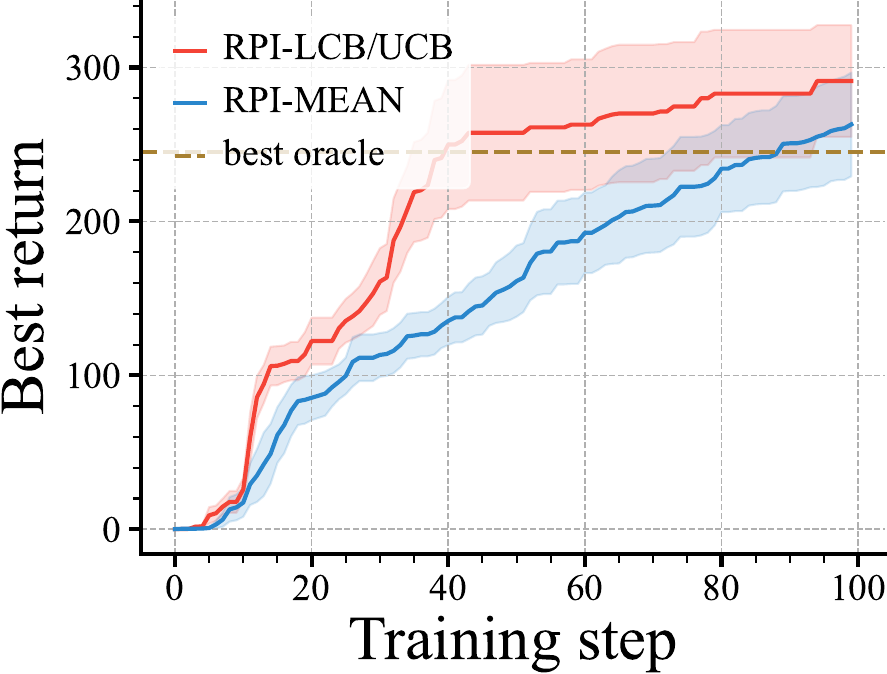}
        \caption{Cheetah}
    \end{subfigure}\hfil
    \begin{subfigure}{.24\textwidth}
        \centering
        \includegraphics[%
        width=3.4cm,  clip={0,0,0,0}]{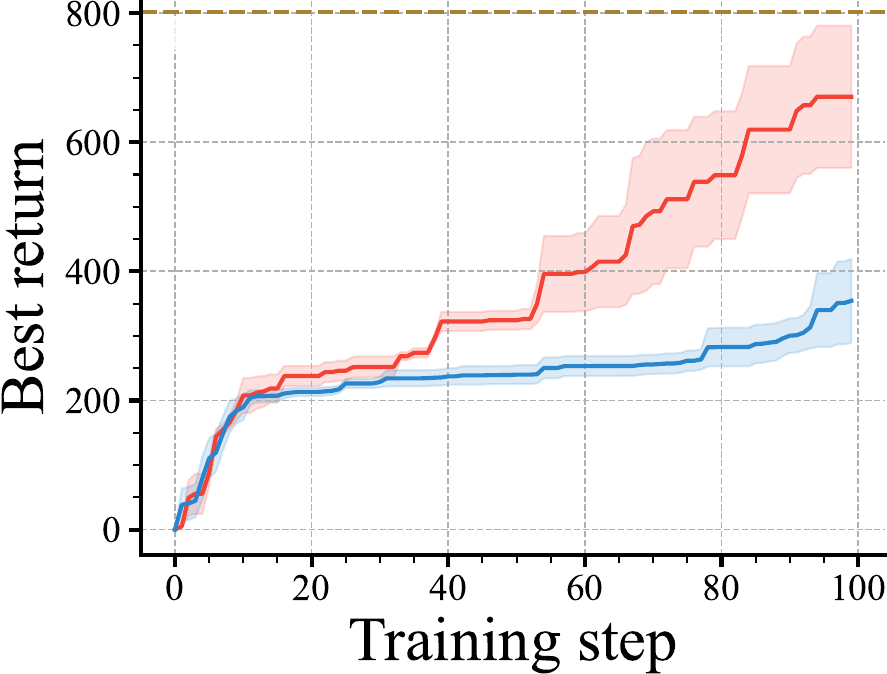}
        \caption{Cartpole}
    \end{subfigure}\hfil
    \begin{subfigure}{.24\textwidth}
        \centering
        \includegraphics[%
        width=3.4cm,  clip={0,0,0,0}]{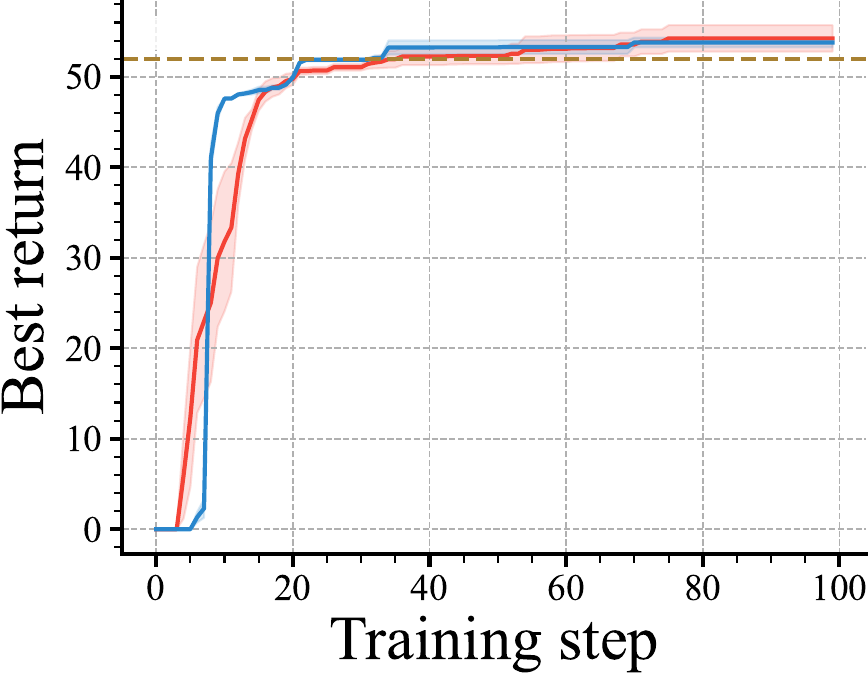}
        \caption{Pendulum}
    \end{subfigure}\hfil
    \begin{subfigure}{.24\textwidth}
        \centering
        \includegraphics[%
        width=3.4cm,  clip={0,0,0,0}]{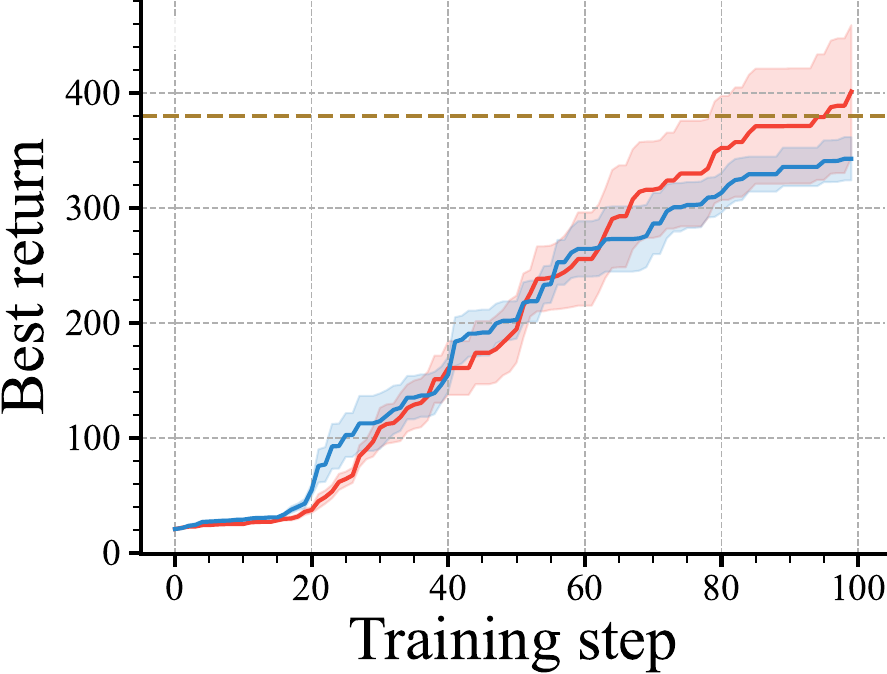}
        \caption{Walker-walk}
    \end{subfigure}\hfil
    \vspace{-2mm}
    \caption{ Experimental results on ablation study on confidence aware UCB/LCB vs MEAN policy selection.}\label{fig:ucbvsmean}
\end{figure*}
\begin{table*}[ht!]
    \centering
    {\scriptsize
    \scalebox{1}{
    \begin{tabular}{l c c c c c c }
        \toprule
        \textbf{Environment}
        & {\makecell[c]{Round}}
        & \makecell[c]{RPI-RAPS(LCB/UCB)}
        & \makecell[c]{RPI-MEAN}
        \\
        \midrule
        \textbf{\makecell[l]{Cheetah}}
        &  \makecell[c]{100}
        &  \makecell[c]{\textbf{291.2} ± 36.3}
        &  \makecell[c]{263.0 ± 33.7} 
        \\
        \textbf{\makecell[l]{Walker-walk}}
        &  \makecell[c]{100}
        &  \makecell[c]{\textbf{402.2} ± 57.7}
        &  \makecell[c]{342.7 ± 18.8}
        \\
        \textbf{\makecell[l]{Pendulum}}
        &  \makecell[c]{100}
        &  \makecell[c]{\textbf{54.3} ± 1.5}
        &  \makecell[c]{53.8 ± 0.5}
        \\
        \textbf{\makecell[l]{Cartpole}}
        &  \makecell[c]{100}
        &  \makecell[c]{\textbf{670.4} ± 110.1}
        &  \makecell[c]{354.3 ± 65.2}
        \\
         \midrule
        \textbf{\makecell[l]{Overall}}
        &  \makecell[c]{100}
        &  \makecell[c]{\textbf{1418.1}}
        &  \makecell[c]{1013.8}
        \\
        \bottomrule
    \end{tabular}}}
    \caption{{Ablation study on confidence aware UCB/LCB vs MEAN policy selection} }
        \label{table:raps-lcb-ucb}   
\end{table*}
We conducted an ablation study that compares RPI-LCB/UCB, which takes uncertainty into account as follows:
\begin{equation*}
 K= \argmax \;(\overline{\hat{V}^1(s)}, \overline{\hat{V}^2(s)}, \ldots, \underline{\hat{V}^{K+1}(s)})   
\end{equation*}

against RPI-MEAN, which does not consider uncertainty as:
\begin{equation*}
K={\argmax} \;(\hat{V^1}(s), \hat{V^2}(s), \ldots, \hat{V}^{K+1}(s)).
\end{equation*}

The experimental results presented in \figref{fig:ucbvsmean} and  \tabref{table:raps-lcb-ucb} demonstrate that the RPI-LCB/UCB  strategy outperforms RPI-MEAN across all benchmarks by an $\textit{overall}$ margin of 40\%. This highlights the significance of incorporating uncertainty in the policy selection strategy.

\ifSubfilesClassLoaded{%
\bibliographystyle{plainnat}
\bibliography{reference}%
}

\end{document}

\end{document}